\documentclass[10pt,twocolumn,letterpaper]{article}

\usepackage[pagenumbers]{cvpr} %

\newcommand{\method}{FLAIR\xspace}

\newcommand{\fullname}{\textbf{F}ine-grained \textbf{La}nguage-informed \textbf{I}mage \textbf{R}epresentations\xspace}
\newcommand{\pooling}{text-conditioned attention pooling}
\newcommand{\myparagraph}[1]{\noindent\textbf{#1}}

\usepackage{tabularray}
\usepackage{booktabs}
\usepackage{multirow}

\definecolor{cvprblue}{rgb}{0.21,0.49,0.74}
\usepackage[pagebackref,breaklinks,colorlinks,allcolors=cvprblue]{hyperref}

\title{FLAIR: VLM with Fine-grained Language-informed Image Representations}

\author{Rui Xiao\textsuperscript{1}, Sanghwan Kim\textsuperscript{1,2,3,4}, Mariana-Iuliana Georgescu\textsuperscript{1,2,3,4}, Zeynep Akata\textsuperscript{1,2,3,4}, Stephan Alaniz\textsuperscript{1,2,3,4}\\[5pt]
\textsuperscript{1}Technical University of Munich \quad \textsuperscript{2}Helmholtz Munich  \quad \textsuperscript{3}MCML \quad \textsuperscript{4}MDSI\\
}

\begin{document}
    \maketitle
    \begin{abstract}
CLIP has shown impressive results in aligning images and texts at scale. However, its ability to capture detailed visual features remains limited because CLIP matches images and texts at a global level.
To address this issue, we propose \method, \fullname, 
an approach that utilizes long and detailed image descriptions to learn localized image embeddings. By sampling diverse sub-captions that describe fine-grained details about an image, we train our vision-language model to produce not only global embeddings but also text-specific image representations. Our model introduces text-conditioned attention pooling on top of local image tokens to produce fine-grained image representations that excel at retrieving detailed image content. We achieve state-of-the-art performance on both, existing multimodal retrieval benchmarks, as well as, our newly introduced fine-grained retrieval task which evaluates vision-language models' ability to retrieve partial image content. Furthermore, our experiments demonstrate the effectiveness of \method trained on 30M image-text pairs in capturing fine-grained visual information, including zero-shot semantic segmentation, outperforming models trained on billions of pairs.
Code is available at \href{https://github.com/ExplainableML/flair}{https://github.com/ExplainableML/flair}.
\end{abstract}
    
    \section{Introduction}
\label{sec:intro}
\begin{figure}[t]
  \centering
   \includegraphics[width=\linewidth]{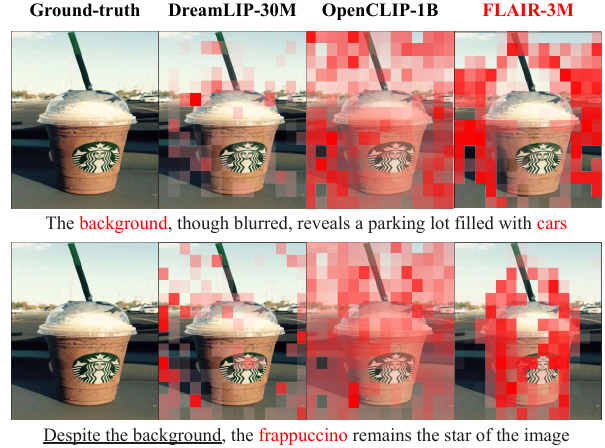}

   \caption{Visualization of the similarity scores between local image tokens and different text queries.
   While previous works~\cite{zheng2025dreamlip, radford2021learning} lack fine-grained alignment,
   \method matches text and image semantics at the token level.
   }
   \label{fig:teaser_small}
   \vspace{-2mm}
\end{figure}

By encoding images and texts into global embeddings, CLIP achieves coarse-grained semantic understanding. However it loses track of the local image details,
e.g. CLIP is not able to perceive the difference between ``background'' and ``frappucino'', resulting in the inability to highlight the relevant regions specified in the text prompt, as illustrated in \cref{fig:teaser_small}.
Recently, it has been shown that CLIP models and other vision language models (VLMs) often lack visual details~\cite{tong2024eyes,tong2024cambrian}.
Thus, our goal is to improve the fine-grained visual understanding of CLIP models which is essential for a wide range of downstream applications, such as image-text retrieval or semantic segmentation.

\begin{figure*}[t]
    \centering
    \includegraphics[width=0.95\linewidth]{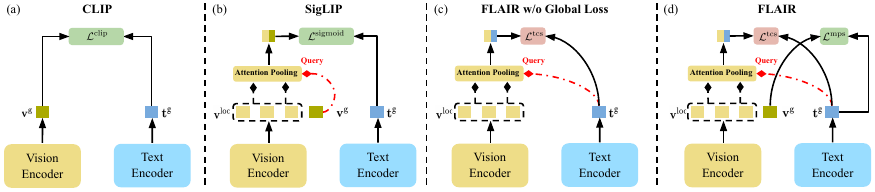}
    \caption{Comparison of text-conditioned attention pooling with previous methods. (a) Vanilla CLIP ($\mathcal{L}^{\text{clip}}$) aligns global image $\mathbf{v}^{\text{g}}$ and text $\mathbf{t}^{\text{g}}$ tokens. (b) SigLIP ($\mathcal{L}^{\text{sigmoid}}$) employs a global learnable image token $\mathbf{v}^{\text{g}}$ as query for a cross-attention to pool the local tokens $\mathbf{v}_{\text{loc}}$. (c) \method ($\mathcal{L}^{\text{tcs}}$) employs text-conditioned attention pooling that leverages $\mathbf{t}^{\text{g}}$ as query, aggregating $\mathbf{v}^{\text{loc}}$ to capture language-informed visual features. (d) \method ($\mathcal{L}^{\text{tcs}} + \mathcal{L}^{\text{mps}}$) adds an extra multi-positive global sigmoid loss to refine global-level image-text alignment.}
    \label{fig:method_overiew}
    \vspace{-2mm}
\end{figure*}

Previous works~\cite{fan2024improving,zheng2025dreamlip} propose to generate detailed descriptions for images to achieve more localized image-text alignment in CLIP models. However, these methods are restricted by the conventional learning mechanism of CLIP, since the detailed text descriptions enhance visual representations indirectly by matching them through the contrastive loss. 
Although DreamLIP~\cite{zheng2025dreamlip} proposed to supervise local image tokens with textual information, we find that, without a careful selection of the negative pairs in the contrastive loss, the VLM does not learn to align the image tokens with semantically matching text, as illustrated in \cref{fig:teaser_small}.

To address these issues, we propose \method, to learn \fullname, where image embeddings are generated by conditioning on a relevant text embedding for a more targeted alignment, instead of an indirect alignment through a global loss function.
To obtain image descriptions with maximum semantic richness, our method leverages long-caption datasets generated by Multimodal Large Language Models (MLLMs). 
These captions provide a rich source of information about specific objects or regions in the image.
Given a long caption, we sample diverse sub-captions, some of which focus on local regions, while others describe the image globally.
Considering that these captions describe the image to varying extents, we design our image encoder to produce text-conditioned image representations. To be specific, we introduce an attention pooling operation that uses the caption as a query to pool relevant image-token embeddings together.

As a result, \method learns fine-grained image embeddings that demonstrate strong performance at retrieving fine-grained visual information.
As shown in Figure \ref{fig:teaser_small}, \method can localize image regions relevant to the fine-grained textual description simply by computing the embedding similarity with respect to the individual image tokens. This is in contrast to previous methods that fail to capture local similarity.
To analyze the text-image retrieval capabilities of our model, we consider three
settings: standard (global) captions, long captions, and a newly proposed fine-grained retrieval setting, where the goal is to match short captions that describe a local region of the image.
Our experimental evaluation on multimodal retrieval and zero-shot semantic segmentation demonstrates that \method, trained on 30M image-text pairs with long synthetic captions, significantly outperforms previous vision-language models trained on billions of image-text pairs. While excelling at fine-grained tasks, \method demonstrates comparable performance on global-level tasks, such as image classification, when trained on the same amount of data.

Our key contributions can be summarized as follows:
1) We propose \method, a model architecture that employs text-conditioned attention pooling to produce fine-grained and localized image embeddings.
2) Building upon long synthetic captions, we introduce a diverse caption sampling strategy to
obtain a rich set of positive and negative image-text pairs facilitating the learning of global and local multimodal relations.
3) Our experimental evaluation on fine-grained downstream tasks shows that \method, trained on 30M samples, outperforms previous models by up to 10.8\% R@1 on coarse-to-fine multimodal retrieval and by up to 11.2\% R@1 on long retrieval tasks. Comparing with CLIP models trained on billions of data, \method achieves an average of 14.4\% increase in mIOU on segmentation tasks.

    \section{Related Works}
\label{sec:related_works}
\myparagraph{Vision-Language Pre-training.} 
CLIP~\cite{radford2021learning} and ALIGN~\cite{jia2021scaling} have scaled up vision-language pre-training datasets to 400M and 1B samples, using a contrastive loss to match global image tokens with global text tokens (\cref{fig:method_overiew} (a)).
However, there is a growing demand for more fine-grained alignment between modalities~\cite{tong2024eyes}.
Several approaches have been proposed to achieve this goal, including token-level alignment~\cite{yao2021filip}, hierarchical alignment from global to local~\cite{gao2022pyramidclip}, soft assignments allowing many-to-many mappings~\cite{gao2024softclip}, and the use of intra-modal contrastive losses~\cite{lee2022uniclip}. CoCa~\cite{yu2022coca} utilizes cross-attention to pool the local image tokens and achieves more refined image-to-text alignment by additionally training with a captioning objective. Along with attention pooling to form the global image embeddings (\cref{fig:method_overiew} (b)), SigLIP \cite{zhai2023sigmoid} replaced the Softmax loss of vision-language pre-training with a Sigmoid-based loss.
Concurrent to our work, Llip~\cite{lavoiemodeling} proposed an architecture incorporating language information into learnable image tokens to form contextualized visual representations. However, Llip~\cite{lavoiemodeling} lacks the pooling of local image tokens (\cref{fig:method_overiew} (c)) and, thus, it does not ensure a fine-grained alignment between modalities.
In contrast, \method leverages diverse and detailed captions with both local and global alignment (\cref{fig:method_overiew} (d)), outperforming previous approaches even when training on a significantly smaller dataset.

\noindent
\myparagraph{Text Augmentation.}
Several works~\cite{fan2024improving,zheng2025dreamlip,wu2024lotlip,zhang2024long} proposed to improve the visual-language alignment through text augmentation. Notably, LaCLIP~\cite{fan2024improving} rewrites captions in large datasets with Large Language Models (LLMs), showing significant performance gains when training on synthetic captions. Similarly, large Vision-Language Models
(VLMs) have been exploited to create synthetic images and captions, augmenting existing datasets~\cite{yang2023alip, hammoud2024synthclip, liu2023mllms}. DreamLIP~\cite{zheng2025dreamlip} re-captions 30 million images from CC3M~\cite{sharma2018conceptual}, CC12M~\cite{changpinyo2021cc12m} and YFCC15M~\cite{cui2022democratizing} with detailed descriptions generated by pre-trained MLLMs. Employing these synthetic captions, several models have been trained to handle long texts going beyond the 77-token limit of CLIP. Long-CLIP~\cite{zhang2024long}, LoTLIP~\cite{wu2024lotlip}, and TULIP~\cite{najdenkoska2024tulip} all leverage synthetic captions to achieve this goal.
Although trained on the same 30M re-captioned images as DreamLIP, \method changes the image-text interaction by directly using text-conditioned attention pooling to aggregate the local image tokens and choosing informative negative pairs in the loss function.  Notably, without modifying the text encoder, the diverse sampling strategy empowers \method to surpass models specialized for long caption retrieval task.

    \begin{figure*}[t]
    \centering
    \includegraphics[width=0.95\linewidth]{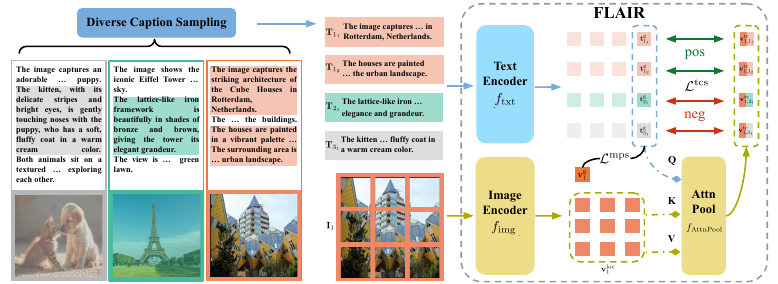}
\caption{Overview of \method; We sample diverse positive and negative captions $\{T_{1_1}...T_{3_1}\}$ for an image $I_1$. $f_{\text{txt}}$ and $f_{\text{img}}$ then produce the global text tokens $\{ \mathbf{t}^{\text{g}}_{1_1} ... \mathbf{t}^{\text{g}}_{3_1} \}$, the global image token $\mathbf{v}^{\text{g}}_1$, and local image tokens $\mathbf{v}^{\text{loc}}_1$. Conditioned on $\{ \mathbf{t}^{\text{g}}_{1_1} ... \mathbf{t}^{\text{g}}_{3_1} \}$, $f_{\text{AttnPool}(.)}$ generates fine-grained text-conditioned image representations  $\{ \mathbf{v}^{\text{tc}}_{1_1} ... \mathbf{v}^{\text{tc}}_{3_1} \}$. The text-conditioned sigmoid loss $\mathcal{L}^{\text{tcs}}$ aligns $\{ \mathbf{t}^{\text{g}}_{1_1} ... \mathbf{t}^{\text{g}}_{3_1} \}$ with $\{ \mathbf{v}^{\text{tc}}_{1_1} ... \mathbf{v}^{\text{tc}}_{3_1} \}$ contrastively, while the multi-positive sigmoid loss $\mathcal{L}^{\text{mps}}$ refines the global alignment between $\mathbf{v}^{\text{g}}_1$ and $\{ \mathbf{t}^{\text{g}}_{1_1} ... \mathbf{t}^{\text{g}}_{3_1} \}$.
}
    \label{fig:pipeline}
    \vspace{-2mm}
\end{figure*}

\section{\method: \fullname}
\label{sec:methodology}

In this section, we present the \method architecture and methodology for language-image pre-training. 
We provide an overview of the main components of \method in~\cref{fig:pipeline}, including the sampling of diverse captions from long synthetic descriptions (\cref{subsec:captions}), the text-conditioned attention pooling of image tokens (\cref{subsec:textcon_pool}), and the local and global loss functions (\cref{subsec:loss_tcs,subsec:loss_mps}).

\subsection{Sampling Diverse Captions}
\label{subsec:captions}
Pre-training data for vision-language models is typically collected by scraping and filtering large amounts of web data, as performed by CC3M~\cite{sharma2018conceptual} or LAION~\cite{schuhmann2021laion,schuhmann2022laion}. While a large amount of image-text pairs helps in discovering a comprehensive set of visual concepts, the text descriptions in these datasets often do not describe the image content in detail. As a result, it is not possible to extract fine-grained concepts in an image, such as scene composition, and small object features. To alleviate this issue, we employ image datasets that are synthetically re-captioned and contain a long and detailed description of each image~\cite{zheng2025dreamlip,singla2024pixels}. A single sentence of these long captions typically describes a particular image detail, e.g., one object, a feature of an object, the background, the image style, or context.

Using these captions, our goal is to align vision and language representations at the fine-grained level of individual caption sentences, while retaining global image understanding. We devise a sampling strategy to cover both local and global captions, and learn their similarity with adaptively pooled image features through a contrastive loss.
Specifically, when constructing a batch of $B$ images from the augmented dataset, we sample $K$ sub-captions from a caption $T_i$ belonging to an image $I_i$. Each sub-caption consists of $s \in \{1,\dots,S\}$ sentences that are either randomly sampled (i.e., independently sampled and concatenated), or extracted as a consecutive sequence of sentences. As a result, a batch contains $B$ images and $B \times K$ texts, where each image is associated with $K$ matching captions. At each iteration, we randomly choose the number of sentences $s$ for every sub-caption, where a lower $s$ result in a more localized caption, while more sentences (a higher $s$)  describe multiple parts of the image, resulting in global descriptions. We provide examples on the original long caption and our sampled captions in Sec.~\ref{sec:example_sampled_captions} in the supplementary.

\subsection{Text-conditioned Attention Pooling}
\label{subsec:textcon_pool}
Having access to a diverse set of captions, some describing local regions of an image and others explaining the global content, motivates creating a model architecture that is capable of adapting to both scenarios. Naively applying a contrastive loss between a global image embedding and the individual text embeddings would collapse the carefully separated information content of our $K$ captions into an averaged image representation.

Instead, we propose to contextualize the image representations with the individual captions, producing a unique image representation for every image-text pair.
We start with the VLM architecture as proposed by~\citet{radford2021learning}, which uses two independent transformer encoders $f_\text{img}$ and $f_\text{txt}$ to project the tokenized image and text samples into per-token embeddings and global embeddings (i.e., $f_{\text{img}}(I) = [\{\mathbf{v}^{(p)}\}_{p=1}^n,\mathbf{v}^{\text{g}}]$ and $f_{\text{txt}}(T) = [\{\mathbf{t}^{(p)}\}_{p=1}^m,\mathbf{t}^{\text{g}}]$), where $n$ is the number of image tokens and $m$ the number of text tokens. For simplicity, we refer to the local image tokens $\{\mathbf{v}^{(p)}\}_{p=1}^n$ as $\mathbf{v}^{\text{loc}} \in \mathbb{R}^{n \times d}$ where $d$ denotes the embedding dimension. 

To effectively contextualize the image representation with semantics from the sampled captions, we introduce an attention pooling layer $f_{\text{AttnPool}}$, that produces a text-conditioned image representation $\mathbf{v}^{\text{tc}}$ from the local image patch embeddings and the global text embedding. We define $\mathbf{v}^{\text{tc}} = f_{\text{AttnPool}}(\mathbf{t}^{\text{g}}, \mathbf{v}^{\text{loc}})$ as follows:
\begin{equation}
    f_{\text{AttnPool}}(\mathbf{t}^{\text{g}}, \mathbf{v}^{\text{loc}}) = \text{softmax} \left( \frac{\mathbf{t}^{\text{g}} W_{\text{q}} (\mathbf{v}^{\text{loc}} W_{\text{k}})^\text{T}}{\sqrt{d}} \right) \mathbf{v}^{\text{loc}} W_{\text{v}}
\end{equation}
where $W_q, W_k, W_v$ are the query, key, and value weight matrices. In other words, we use the global text embeddings of a caption as a query to pool the local image embeddings creating a text-conditioned image representation $\mathbf{v}^{\text{tc}}$. In practice, we use a multi-head attention layer. We append an empty token (zero vector) to $\mathbf{v}^{\text{loc}}$ to allow $\mathbf{t}^{\text{g}}$ to attend to the empty token when $\mathbf{t}^{\text{g}}$ and $\mathbf{v}^{\text{loc}}$ are not semantically related.

\noindent
\myparagraph{Choosing Negative Pairs.}
With text-conditioned attention pooling, \method produces a different image representation for every image-text pair. However, to learn semantically rich and nuanced image representations, we need to carefully define the positive and negative pairs for vision-language pre-training. To simplify notation, we assume a single caption per image (i.e., $K=1$). Let $\mathbf{t}^{\text{g}}_{i}$ be the caption of the $i$-th image in the batch, and $\mathbf{v}^{\text{tc}}_{i,j}$ be the image embedding from the $i$-th image conditioned on the caption of image $j$. For the explanation in this section only, we enforce $i \neq j$. 
In the context of contrastive learning, image-text pairs where image, caption, and condition come from the same sample are considered positive pairs. In other words, $\langle \mathbf{v}^{\text{tc}}_{i,i}, \mathbf{t}^{\text{g}}_{i} \rangle$ is maximized during training, where $\langle.,.\rangle$ denotes the cosine similarity. For negative pairs, the text condition introduces multiple options. DreamLIP~\cite{zheng2025dreamlip} proposed a loss that uses negatives defined as $\langle \mathbf{v}^{\text{tc}}_{i,j}, \mathbf{t}^{\text{g}}_{i} \rangle$ in our formulation.
However, this allows to solve the contrastive objective by comparing the text condition with the text embedding, ignoring image information and creating an undesired shortcut.
To overcome this problem, we instead propose to adopt $\langle \mathbf{v}^{\text{tc}}_{i,j}, \mathbf{t}^{\text{g}}_{j} \rangle$ as negative pairs. This ensures that image and text representations are contrasted meaningfully, i.e., neither image nor text information can be ignored.
This definition generalizes to multiple captions per image (i.e., $K>1$), where each sub-caption of the same image is considered a positive match, and negative otherwise.
In this case, we can write the full notation as $\langle \mathbf{v}^{\text{tc}}_{i,i_k}, \mathbf{t}^{\text{g}}_{i_k} \rangle$ for positive pairs and $\langle \mathbf{v}^{\text{tc}}_{i,j_k}, \mathbf{t}^{\text{g}}_{j_k} \rangle$ for negative pairs, where $k$ is the sub-caption index of the $j$-th image. Consequently, these positive and negative pairs allow \method to learn text-aware image representations. Extended analysis is provided in Sec.~\ref{sec:supp_negatives} in the supplementary.

\subsection{Text-conditioned Sigmoid Loss}
\label{subsec:loss_tcs}After constructing the positives and negatives pairs and applying $f_{\text{AttnPool}}(.)$, we adopt a contrastive loss based on the sigmoid function as proposed by SigLIP~\cite{zhai2023sigmoid}. It is preferred over the InfoNCE loss~\cite{oord2018representation}, as it enables multiple positive pairs in the same batch, and is more efficient for multi-GPU training.
Accordingly, we define our text-conditioned sigmoid loss as
\begin{equation}
    \mathcal{L}_{i,j,k}^{\text{tcs}} = \frac{1}{1 + e^{y_{i,j} \left( - t \langle \mathbf{v}^{\text{tc}}_{i,j_k}, \mathbf{t}^{\text{g}}_{j_k} \rangle + b \right)}}
\end{equation}
where $t$ is a learnable temperature, $b$ is a learnable bias, and $\langle \cdot, \cdot \rangle$ is the cosine similarity. $y_{i,j}$ is $+1$ for positive pairs when $i=j$ for all $k \in [1,\dots,K]$, and $-1$ for negative pairs otherwise.
Since every batch contains $B$ images and $BK$ captions, we reduce the compute and memory usage of $\mathcal{L}_{i,j,k}^{\text{tcs}}$ by considering all $K$ positive pairs, but only $B-1$ negative pairs per image, i.e., 1 out of $K$ captions for every negative. Therefore, we compute the similarity of $B \times (K+B-1)$ pairs, instead of $B \times BK$ pairs for every batch.

$\mathcal{L}^{\text{tcs}}$ aligns the text-conditioned image embedding with the corresponding text embedding. Intuitively, this allows the image encoder to store semantic information locally in each token and pool the relevant tokens based on the text query producing context-aware representations.
Our main experiments demonstrate that the text-conditioned image embeddings contribute significantly to fine-grained image-text alignment, providing the majority of the performance improvement in zero-shot semantic segmentation.

\subsection{Multi-positive Sigmoid Loss}
\label{subsec:loss_mps}
We find that \method can be trained exclusively with the $\mathcal{L}^{\text{tcs}}$ loss. At the same time, it proves beneficial to additionally match the global image embedding $\mathbf{v}^{\text{g}}$ with every sub-caption, to also learn a coarse alignment.
Following previous works~\cite{fan2024improving, zheng2025dreamlip, li2021supervision}, we introduce a multi-positive loss to align the global image embedding $\mathbf{v}^{\text{g}}$ with the text embedding $\mathbf{t}^{\text{g}}$ of every sub-caption. Different from previous works, we employ the contrastive sigmoid loss to handle multiple positive captions per image in a more natural way. 
Our multi-positive sigmoid loss is defined as
\begin{equation}
    \mathcal{L}_{i,j,k}^{\text{mps}} = \frac{1}{1 + e^{y_{i,j} \left( - t \langle \mathbf{v}^{\text{g}}_{i}, \mathbf{t}^{\text{g}}_{j_k} \rangle + b \right)}}
\end{equation}
where $y_{i,j}$ is $+1$ for all $k \in [1,\dots,K]$ positive pairs with $i=j$, and $-1$ for negative pairs otherwise. Equivalently to $\mathcal{L}^{\text{tcs}}$, we use all $K$ positive pairs per image and $1$ caption from every $K$ negative sub-captions per match.

Since $\mathcal{L}^{\text{mps}}$ mainly optimizes the global image and text embeddings, it is beneficial for coarse-grained tasks. We empirically find that combining \(\mathcal{L}^{\text{mps}}\) with \(\mathcal{L}^{\text{tcs}}\) consistently improves performance across all tasks, particularly in zero-shot image classification, where global-level alignment is crucial.
Our final loss $\mathcal{L}$ is an average of both losses and it is defined by
\begin{equation} 
    \mathcal{L} = \frac{1}{2}(\mathcal{L}^{\text{tcs}}  + \mathcal{L}^{\text{mps}}).
\label{eq:l_total}
\end{equation}

    \begin{table*}[t!]
	\centering\scriptsize
	\SetTblrInner{rowsep=1.2pt}
	\SetTblrInner{colsep=2pt}
	\resizebox{\linewidth}{!}{
		\begin{tblr}{
			cells={halign=c,valign=m},
			column{1}={halign=l},
			hline{2,23}={1-18}{1pt},
			hline{5,9,13,17}={1-18}{},
			hline{3}={3-6}{leftpos = -1, rightpos = -1, endpos},
			hline{3}={7-10}{leftpos = -1, rightpos = -1, endpos},
			hline{3}={11-14}{leftpos = -1, rightpos = -1, endpos},
			hline{3}={15-18}{leftpos = -1, rightpos = -1, endpos},
                vline{11}={2-23}{},
					cell{2}{1}={r=3}{},
					cell{2}{2}={r=3}{},
					cell{1}{3}={c=8}{},
					cell{1}{11}={c=8}{},
					cell{2}{3}={c=4}{},
					cell{2}{7}={c=4}{},
					cell{2}{11}={c=4}{},
					cell{2}{15}={c=4}{},
					cell{3}{3}={c=2}{},
					cell{3}{5}={c=2}{},
					cell{3}{7}={c=2}{},
					cell{3}{9}={c=2}{},
					cell{3}{11}={c=2}{},
					cell{3}{13}={c=2}{},
					cell{3}{15}={c=2}{},
					cell{3}{17}={c=2}{},
					cell{5}{1}={r=4}{},
					cell{9}{1}={r=4}{},
					cell{13}{1}={r=4}{},
					cell{17}{1}={r=6}{},
				}

				            &                                         & \SetCell[c=8]{c} \bf Standard Retrieval &          &                      &          &                            &           &                      &          & \SetCell[c=8]{c} \bf Fine-grained Retrieval &           &                      &          & \SetCell[c=4]{c} IIW &          &                      &          \\
			Setting         & Method                                  & \SetCell[c=4]{c} MSCOCO                 &          &                      &          & \SetCell[c=4]{c} Flickr30k &           &                      &          & \SetCell[c=4]{c} DOCCI-FG                      &           &                      &          & \SetCell[c=4]{c} IIW-FG &          &                      &          \\
			                &                                         & \SetCell[c=2]{c} T2I                    &          & \SetCell[c=2]{c} I2T &          & \SetCell[c=2]{c} T2I       &           & \SetCell[c=2]{c} I2T &          & \SetCell[c=2]{c} T2I                        &           & \SetCell[c=2]{c} I2T &          & \SetCell[c=2]{c} T2I &          & \SetCell[c=2]{c} I2T &          \\
			                &                                         & R@1                                     & R@5      & R@1                  & R@5      & R@1                        & R@5       & R@1                  & R@5      & R@1                                         & R@5       & R@1                  & R@5      & R@1                  & R@5      & R@1                  & R@5      \\
			CC3M-recap       & CLIP~\cite{radford2021learning}         & 27.0                                    & 52.6     & 38.9                 & 66.1     & 49.9                       & 77.0      & 67.8                 & 88.5     & 10.3                                        & 23.4      & 25.0                 & 50.9     & 24.4                 & 45.0     & 61.1                 & 85.6     \\
			                & SigLIP~\cite{zhai2023sigmoid}           & 28.3                                    & 54.4     & 40.1                 & 67.5     & 53.2                       & 78.5      & 69.9                 & 90.4     & 10.4                                        & 23.8      & 24.9                 & 50.6     & 24.7                 & 45.1     & 62.6                 & 86.9     \\
			                & DreamLIP~\cite{zheng2025dreamlip}       & 29.8                                    & 55.4     & 40.8                 & 68.4     & 53.6                       & 78.4      & 69.2                 & 91.5     & 10.3                                        & 22.8      & 23.3                 & 47.5     & 22.7                 & 41.6     & 59.2                 & 83.3     \\
			                & \method                                 & \bf 37.7                                & \bf 64.5 & \bf 51.6             & \bf 77.2 & \bf 65.7                   & \bf 86.8  & \bf 78.7             & \bf 95.2 & \bf15.1                                     & \bf30.9   & \bf35.7              & \bf63.5  & \bf30.5              & \bf52.3  & \bf70.6              & \bf90.9  \\
			CC12M-recap      & CLIP~\cite{radford2021learning}         & 39.8                                    & 66.4     & 56.2                 & 80.5     & 67.0                       & 87.7      & 81.7                 & 96.5     & 16.0                                        & 31.6      & 39.5                 & 66.5     & 31.8                 & 52.9     & 76.4                 & 93.6     \\
			                & SigLIP~\cite{zhai2023sigmoid}           & 40.4                                    & 67.0     & 55.3                 & 79.7     & 66.7                       & 88.1      & 82.5                 & 96.1     & 16.2                                        & 31.9      & 40.0                 & 66.8     & 31.9                 & 53.2     & 78.4                 & 94.3     \\
			                & DreamLIP~\cite{zheng2025dreamlip}       & 40.6                                    & 66.5     & 54.0                 & 78.3     & 68.3                       & 89.3      & 84.1                 & 97.8     & 17.2                                        & 33.0      & 41.6                 & 68.5     & 31.9                 & 52.1     & 77.8                 & 94.9     \\
			                & \method                                 & \bf 47.8                                & \bf 73.5 & \bf 64.1             & \bf 85.0 & \bf 75.4                   & \bf 92.15 & \bf 90.8             & \bf 98.4 & \bf21.4                                     & \bf38.8   & \bf50.4              & \bf76.7  & \bf38.7              & \bf59.9  & \bf83.8              & \bf96.9  \\
			YFCC15M-recap    & CLIP~\cite{radford2021learning}         & 44.7                                    & 71.2     & 61.0                 & 85.0     & 72.3                       & 90.8      & 89.1                 & 97.6     & 18.1                                        & 35.3      & 43.1                 & 71.9     & 34.4                 & 56.5     & 81.4                 & 96.7     \\
			                & SigLIP~\cite{zhai2023sigmoid}           & 46.6                                    & 72.8     & 62.6                 & 85.3     & 73.6                       & 92.1      & 90.0                 & 97.6     & 18.9                                        & 35.8      & 46.3                 & 74.8     & 35.5                 & 56.6     & 84.3                 & 96.2     \\
			                & DreamLIP~\cite{zheng2025dreamlip}       & 42.4                                    & 68.5     & 57.0                 & 81.0     & 70.0                       & 89.2      & 87.3                 & 98.1     & 17.3                                        & 33.6      & 41.4                 & 69.8     & 32.0                 & 53.0     & 76.1                 & 95.4     \\
			                & \method                                 & \bf 51.2                                & \bf 76.0 & \bf 67.3             & \bf 88.1 & \bf 79.2                   & \bf 94.2  & \bf 93.3             & \bf 99.1 & \bf 23.0                                    & \bf  41.2 & \bf 53.7             & \bf 79.7 & \bf 39.5             & \bf 62.1 & \bf 85.5             & \bf 96.4 \\
                            
			SOTA Comparison & OpenCLIP (2B)~\cite{cherti2023reproducible}     &              41.7                     &   67.1   &      59.3       &  82.4   &        71.9                 &    90.4  &         87.5      &    97.7    &                 17.4                    &   31.9   &       49.7       &   75.9  &         30.6         &    48.4  &          84.1       &  95.4   \\
			                & SigLIP (10B)~\cite{zhai2023sigmoid}       & 47.2                                    & 72.1     & 65.5                 & 86.2     & 75.6                       & 92.8      & 89.1                 & 98.6     & 20.6                                        & 35.9      & 57.5                 & 82.1     & 33.8                 & 53.0     & 83.7                 & 97.7     \\
			                & MetaCLIP (2.5B)~\cite{xu2023demystifying} & 41.4                                    & 67.2     & 59.4                 & 80.6     & 76.2                       & 90.7      & 85.9                 & 97.3     & -                                           & -         & -                    & -        & -                    & -        & -                    & -        \\
			                & Llip (2.5B)~\cite{lavoiemodeling}     & 45.6                                    & 70.8     & 63.4                 & 84.3     & 75.1                       & 92.8      & 90.1                 & 98.5     & -                                           & -         & -                    & -        & -                    & -        & -                    & -        \\
			                & DreamLIP (30M)~\cite{zheng2025dreamlip}   & 44.8                                    & 69.8     & 62.3                 & 84.5     & 73.3                       & 91.8      & 89.9                 & 99.0     & 21.6                                        & 39.3      & 51.2                 & 78.3     & 37.5                 & 58.6     & 85.3                 & 97.4     \\
			                & \method (30M)                             & \bf 53.3                                & \bf 77.5 & \bf 68.0             & \bf 87.8 & \bf 81.1                   & \bf 94.9  & \bf 94.7             & \bf 99.3 & \bf 25.0                                    & \bf 43.8  & \bf 59.0             & \bf 84.1 & \bf 41.7             & \bf 63.4 & \bf 91.5             & \bf 98.9 \\
		\end{tblr}}
	\caption{Zero-shot image-text retrieval on validation splits for standard benchmarks (Flickr30k~\cite{plummer2015flickr30k} and MSCOCO~\cite{lin2014microsoft}) and our introduced fine-grained retrieval setting (sentence-level on DOCCI~\cite{onoe2024docci} and IIW~\cite{garg2024imageinwords}). Except for ``SOTA Comparison'', all models are pre-trained on CC3M-recap, CC12M-recap, YFCC15M-recap, under the same training configurations. All models use ViT-B/16 as the vision encoder.
    }
	\label{tab:retrieval_full}
\end{table*}

\section{Experiments}
We present our experimental evaluation of \method on the three image-text retrieval settings: standard (\cref{sec:retrieval_standard}), fine-grained (\cref{sec:retrieval_finegrained}), and long (\cref{sec:retrieval_long}).
In addition, we conduct experiments on zero-shot semantic segmentation (\cref{sec:segmentation}) and image classification (\cref{sec:classification}), qualitatively evaluate the attention maps of \method (\cref{sec:qualitative}), and ablate important model components (\cref{sec:ablation_study}).

\subsection{Experimental Setup}
\myparagraph{Pre-training Datasets.}
To learn fine-grained image embeddings from descriptive local captions, we pre-train \method on DreamLIP's~\cite{zheng2025dreamlip} re-captioned
datasets, which we refer to
as CC3M-recap, CC12M-recap, and YFCC15M-recap. Following DreamLIP, we also merged these three datasets into a combined set of 30M samples.

\noindent
\myparagraph{Implementation Details.} Our model is based on the OpenCLIP~\cite{cherti2023reproducible} code implementation, adopting their default settings. We use ViT-B/16 as the vision encoder, with the default pre-processing:
images are resized to $224\times224$ pixels, and text sequences are tokenized to a maximum length of 77 tokens. For direct comparison with DreamLIP, we follow their training configuration and caption pre-processing, splitting the MLLM-generated and original captions into individual sentences.
To obtain diverse training captions, we sample $K=8$ captions per image, with each caption randomly merging 1 to 3 sentences (i.e., $S=3$). To maximize sampling variability while retaining context, we randomly construct our sub-caption by either sampling consecutive sentences or merging sentences from random positions in the original text.
For fair comparison, we reproduce CLIP~\cite{radford2021learning} and SigLIP~\cite{zhai2023sigmoid} on all re-captioned datasets under identical training configurations
as \method.
Further details are available in Sec.~\ref{sec:implementation_details} in the supplementary.

\noindent
\myparagraph{Inference with \method.}
To utilize the fine-grained embeddings from \method for image-to-text (I2T) retrieval, each image $i$ is first conditioned on all $j$ texts to generate the conditioned embeddings \( \mathbf{v}^{\text{tc}}_{i,j} \). Then we compute the similarity scores between the conditioned embeddings and each text embedding
($\langle \mathbf{v}^{\text{tc}}_{i,j}, \mathbf{t}^{\text{g}}_{j} \rangle$)
to obtain Recall@K from the top-K retrieval items.
The text-to-image (T2I) retrieval score matrix is the transpose of the image-to-text retrieval matrix.

\begin{table}[t]
    \centering\scriptsize
	\SetTblrInner{rowsep=1.2pt}
	\SetTblrInner{colsep=2pt}
    \resizebox{\columnwidth}{!}{
    \begin{tblr}{
        cells={halign=c,valign=m},   %
        hline{1,13}={1.0pt},       %
        hline{3,9}={},       %
        cell{1}{3,5,7,9}={c=2}{c},   %
        cell{1}{1}={r=2}{c},          %
        cell{1}{2}={r=2}{c},          %
    }
        Method & Data & DCI && SV-1k && SV-10k && Urban-1k & \\
                 &    & I2T & T2I & I2T & T2I & I2T & T2I & I2T & T2I \\
        OpenCLIP~\cite{radford2021learning} & 2B & 56.0 &  55.4  & 90.3 & 87.7 & 69.6 & 66.8 & 69.5 & 65.8 \\
        LiT~\cite{zhai2022lit} & 100M & 41.7 & 40.9 & 86.0 & 80.0 & 61.4 & 50.6 & - & - \\
        ALIGN~\cite{jia2021scaling} & 700M & 56.5 & 57.4 & 86.3 & 85.3 & 65.1 & 62.7 & - & - \\
        SigLIP~\cite{zhai2023sigmoid} & 10B & 57.7 & 56.2 & 85.8 & 83.4 & 83.4 & 63.0 & 62.7 & 62.1 \\
        Long-CLIP~\cite{zhang2024long} & 400M & 47.4 & 44.1 & 90.6 & 87.4 & 73.1 & 62.0 & 78.9 & 79.5 \\
        LoTLIP~\cite{wu2024lotlip} & 100M & \bf 62.1 & 61.0 & 95.5 & 86.8 & 81.4 & 83.7 & - & - \\
        \method & 3M & 47.3 & 50.5 & 91.0 & 89.7 & 72.0 & 70.6 & 63.5 & 69.5 \\
        \method & 12M & 55.5 & 60.8 & 96.1 & 95.1 & 85.0 & 83.4 & 74.6 & 80.6 \\
        \method & 15M & 54.9 & \underline{62.4} & \underline{97.4} & \underline{96.7} & \underline{88.8} & \underline{86.8} &  \underline{82.4} &  \underline{86.6}\\
        \method & 30M & \underline{61.3} & \bf 66.2 & \bf 98.5 & \bf 98.0 & \bf 90.3 & \bf 89.4 & \bf 83.6 &  \bf87.7 \\
    \end{tblr}}
    \caption{Zero-shot long text-image retrieval tasks. I2T and T2I indicate the R@1 score on image-to-text and text-to-image retrieval, respectively. The best results are \textbf{bold}, second-best are \underline{underlined}. All models use ViT-B/16 as vision encoder.
    }
    \label{tab:long_retrieval_small}
\end{table}

\begin{table*}[tb]
\parbox{.4\linewidth}{
    \centering\scriptsize
	\SetTblrInner{rowsep=1.2pt}
	\SetTblrInner{colsep=1.4pt}
    \resizebox{.84\columnwidth}{!}{
    \begin{tblr}{
        cells={halign=c,valign=m},   %
        hline{1,12}={1.0pt},       %
        hline{2,5,8}={},       %
        cell{5}{2}={r=3}{},   %
    }
        Method &   \rotatebox[origin=lb]{90}{\smash{ Data Size  }} &   \rotatebox[origin=lb]{90}{\smash{ VOC20  }}&  \rotatebox[origin=lb]{90}{\smash{ Cityscapes }}  &   \rotatebox[origin=lb]{90}{\smash{ Context59 }}  &   \rotatebox[origin=lb]{90}{\smash{ ADE20K }} & \rotatebox[origin=lb]{90}{\smash{ COCO-Stuff }} &  \rotatebox[origin=lb]{90}{\smash{Average }} \\
        CLIP~\cite{radford2021learning} & 400M & 41.8  & 5.5  &  9.2  &  3.2 & 4.4  &  12.8 \\
        OpenCLIP~\cite{cherti2023reproducible} & 2B &  47.2 &  5.1 &  9.0  &  2.9 & 5.0  &  13.9  \\
        MetaCLIP~\cite{xu2023demystifying} & 2.5B & 35.4  &  5.0 & 8.1  &  2.2 & 4.3  & 11.0 \\
        CLIP~\cite{zhai2023sigmoid} & 30M & 11.3  &   5.0  & 4.5    & 1.3  &  2.8 &  5.0\\
        SigLIP~\cite{zhai2023sigmoid} &  & 14.5  &  5.5  &  5.8  &  2.2 & 3.8  &  6.4 \\
        DreamLIP~\cite{zheng2025dreamlip} &  &  1.8 & 0.9  &  0.4 & 0.1  & 0.1  &  0.7 \\	
        \method & 3M &  60.9 & \bf 20.6  & \bf 23.8 & 13.2 &  13.1 & 26.3  \\
        \method & 12M &  69.7 &  20.1 &  22.9 & \bf 13.3 & \bf 15.4  & \bf 28.3 \\
        \method & 15M & 66.7 & 16.5 & 17.4 & 9.1 & 13.6  &  24.7 \\
        \method & 30M & \bf 73.0  & 13.6  & 18.6  &  10.4 & 13.3  &   25.8\\  	
    \end{tblr}}
    \caption{Mean intersection over union (mIoU) for zero-shot semantic segmentation on the VOC20~\cite{everingham2015pascal}, Cityscapes~\cite{cordts2016cityscapes}, Context59~\cite{mottaghi2014role}, ADE20K~\cite{zhou2019semantic}, and COCO-Stuff~\cite{caesar2018coco} datasets.
    All models employ ViT-B/16 as vision encoder. 
    }
    \label{tab:ss}
}
\hfill
\parbox{.58\linewidth}{
	\centering\scriptsize
	\SetTblrInner{rowsep=1.2pt}
	\SetTblrInner{colsep=1.4pt}
	\resizebox{1.2\columnwidth}{!}{
		\begin{tblr}{
			cells={halign=c,valign=m},   %
			hline{1,15}={1.0pt},       %
					hline{2}={},
					hline{8}={},
					hline{12}={},
					cell{2}{2}={r=6}{},
					cell{8}{2}={r=4}{},
				}
			Method                                 & \rotatebox[origin=lb]{90}{\smash{ Data Size  }} & \rotatebox[origin=lb]{90}{\smash{ Food-101 }} & \rotatebox[origin=lb]{90}{\smash{ CIFAR-10 }} & \rotatebox[origin=lb]{90}{\smash{ CIFAR-100 }} & \rotatebox[origin=lb]{90}{\smash{ SUN397 }} & \rotatebox[origin=lb]{90}{\smash{ Cars }} & \rotatebox[origin=lb]{90}{\smash{ Aircraft }} & \rotatebox[origin=lb]{90}{\smash{ DTD }} & \rotatebox[origin=lb]{90}{\smash{ Pets }} & \rotatebox[origin=lb]{90}{\smash{ Caltech-101 }} & \rotatebox[origin=lb]{90}{\smash{ Flowers }} & \rotatebox[origin=lb]{90}{\smash{ ImageNet }} & \rotatebox[origin=lb]{90}{\smash{ Average}} \\
			LaCLIP~\cite{fan2024improving}         & 3M   & 14.2                                         & 57.1                                         & 27.5                                          & 35.1                                       & 1.6                                      & 1.6                                          & 16.6                                    & 15.6                                     & 52.7                                            & 14.7                                        & 21.5                                         & 23.5                                     \\
			MLLM-A~\cite{liu2023mllms}             &      & 18.7                                         & 58.4                                         & 32.4                                          & 43.8                                       & 3.9                                      & 1.5                                          & 20.2                                    & 32.1                                     & 63.5                                            & 17.5                                        & 25.0                                         & 28.8                                     \\
			CLIP~\cite{radford2021learning}        &      & 17.9                                         & 75.0                                         & 40.8                                          & 43.1                                       & 2.6                                      & 1.0                                          & 15.3                                    & 22.1                                     & 68.9                                            & 12.6                                        & 23.8                                         & 29.4                                     \\
			SigLIP~\cite{zhai2023sigmoid}          &      & 18.4                                         & \underline{76.4}                                         & 41.9                                          & 46.9                                       & 3.0                                      & 1.4                                          & 17.6                                    & 20.6                                     & 70.4                                            & 10.8                                        & 25.4                                         & 30.3                                     \\

			DreamLIP~\cite{zheng2025dreamlip}      &      & \underline{23.1}                             & 75.9                                         & \underline{44.2}                              & 46.6                                       & \underline{3.4}                          & \underline{1.6}                              & \underline{19.0}                        & \underline{27.4}                         & 66.1                                            & \underline{16.0}                            & \underline{30.1}                             & \underline{32.1}                         \\
			\method                                &      & \textbf{24.2}                                & \textbf{82.0}                                & \textbf{51.5}                                 & \textbf{53.8}                              & \textbf{3.7}                             & \textbf{1.7}                                 & \textbf{23.9}                           & \textbf{34.2}                            & \textbf{70.1}                                   & \textbf{19.1}                               & \textbf{33.8}                                & \bf 36.2                                 \\
			CLIP~\cite{radford2021learning}        & 30M  & 61.3                                         & 92.2                                         & 66.9                                          & 62.2                                       & 19.3                                     & 5.7                                          & 30.9                                    & 49.3                                     & 83.7                                            & 43.4                                        & 50.0                                         & 51.4                                     \\
			SigLIP~\cite{zhai2023sigmoid}          &      & 64.2                                         & 91.0                                         & 67.6                                          & \underline{64.0}                           & 22.0                                     & 5.7                                          & 33.5                                    & 53.3                                     & 84.3                                            & 43.6                                        & 51.0                                         & 52.7                                     \\
			DreamLIP~\cite{zheng2025dreamlip}      &      & \textbf{75.4}                                & \underline{92.3}                             & \textbf{70.7}                                 & 63.7                                       & \underline{22.7}                         & \textbf{7.9}                                 & \underline{33.9}                        & \textbf{64.1}                            & \textbf{88.0}                                   & \textbf{51.1}                               & \textbf{58.1}                                & \bf 57.0 \\
			\method                                &      & \underline{72.5}                             & \textbf{93.1}                                & \underline{69.6}                              & \textbf{66.9}                              & \textbf{31.1}                            & \underline{7.2}                              & \textbf{37.3}                           & \underline{55.6}                         & \underline{86.5}                                & \underline{48.4}                            & \underline{56.6}                             & \underline{56.8}                         \\
            
			OpenCLIP~\cite{cherti2023reproducible} & 2B & \underline{86.2}                                         & 94.8                                         & 76.5                                          & \underline{70.0}                                       & 87.4                                     & 25.8                                         & 54.9                                    & 89.5                                     & 93.2                                            & 69.8                                        &   70.2                                         &       74.4                              \\
			MetaCLIP~\cite{xu2023demystifying}     & 2.5B & 88.3                                         & \textbf{95.7}                                         & \underline{79.0}                                          & 68.5                                     & \underline{82.9}                                     & \underline{30.3}                                         & \underline{62.1}                                    & \textbf{91.7}                                     & \underline{93.3}                                            & \underline{73.9}                                       & \underline{72.1}                                         & \underline{76.2}                                     \\
			Llip~\cite{lavoiemodeling}         & 2.5B & \textbf{89.0}                                & \textbf{95.7}                                & \textbf{81.4}                                 & \textbf{70.9}                              & \textbf{88.2}                            & \textbf{41.5}                                & \textbf{63.7}                           & \textbf{93.5}                            & \underline{94.7}                                   & \textbf{74.9}                               & \textbf{75.3}                                & \textbf{79.0}                            \\
		\end{tblr}}
	\caption{Top-1 accuracy for zero-shot classification on: 
        Food-101~\citep{bossardFood101MiningDiscriminative2014},
        CIFAR-10 \& CIFAR-100~\citep{krizhevskyLearningMultipleLayers},
        SUN397~\citep{xiaoSUNDatabaseLargescale2010}, 
        Cars~\citep{krauseCollectingLargeScaleDataset},
        Aircraft~\citep{majiFineGrainedVisualClassification2013}, 
        DTD~\cite{cimpoi2014describing},
        Pets~\citep{parkhiCatsDogs2012}, 
        Caltech-101~\citep{fei2004learning},
        Flowers~\citep{nilsbackAutomatedFlowerClassification2008}, 
        ImageNet~\citep{deng2009imagenet}.
    All models use ViT-B/16 as vision encoder.
    The best and second-best results are \textbf{bold} and \underline{underlined}. 
    }
	\label{tab:zs_cla}

}
\vspace{-2mm}
\end{table*}

\subsection{Standard Zero-shot Image-text Retrieval}
\label{sec:retrieval_standard}
As a standard assessment of image-text alignment, we follow prior works~\cite{zheng2025dreamlip, yang2023alip, lavoiemodeling} to evaluate image-text retrieval on the validation splits of MSCOCO~\cite{lin2014microsoft} and Flickr30K~\cite{plummer2015flickr30k}, where each image is typically paired with five global captions.

\noindent
\myparagraph{Results.}
We report the results on the standard retrieval task in the left side of~\cref{tab:retrieval_full}.
\method outperforms the three baselines, CLIP, SigLIP, and DreamLIP on all pre-training datasets by a large margin. Comparing models trained on CC3M-recap, \method surpasses DreamLIP in the retrieval task, obtaining higher R@1 scores on both COCO (T2I: +7.9\%, I2T: +10.8\%) and Flickr30k (T2I: +12.1\%, I2T: +9.5\%) datasets.
When including SOTA models, \method trained on CC12M-recap obtains a similar performance to SigLIP trained on 10B samples, and surpasses it significantly once we move to larger datasets with YFCC15M-recap and the merged 30M samples. \method-30M (vs. SigLIP-10B) achieves
81.1\% (vs. 75.6\%) T2I, 94.7\% (vs. 89.1\%) I2T on Flickr30k and is similarly better on COCO.
We also notice that CLIP and SigLIP trained on YFCC15M-recap can match or surpass their counterparts trained on billions of data samples.
This suggests two key insights: 1) text augmentations from long synthetic captions empowers VLMs with better retrieval capability, and 2) \method with \pooling\ generates more targeted image embeddings for retrieval, maximizing the benefits from text augmentations, and resulting in a significant improvement with much less image data.

\subsection{Fine-grained Zero-shot Image-text Retrieval}
\label{sec:retrieval_finegrained}
Standard retrieval tasks do not fully capture a model’s ability to align detailed descriptions with images. To address this, we introduce a fine-grained retrieval task aimed at evaluating how well a model can associate an image with fine-grained captions. Our benchmark is constructed as follows: 1) We use the recently released densely-captioned datasets DOCCI~\cite{onoe2024docci} and IIW~\cite{garg2024imageinwords}.
Due to the careful human annotation process, their long captions are free from hallucinations; 2) For each test image in DOCCI and IIW, we split the long captions into individual sentences, yielding an average of 7.1 captions per image in DOCCI and 10.1 in IIW.
As shown in \cref{fig:docci_example}, each caption focuses on a specific local part of the image, making both T2I and I2T tasks significantly more challenging than standard retrieval.
We refer to our split datasets as DOCCI-FG and IIW-FG.

\noindent
\myparagraph{Results.}
The results obtained on  DOCCI-FG and IIW-FG are reported in the right side of~\cref{tab:retrieval_full}.
The difficulty of this task is apparent by the significantly lower text-to-image (T2I) retrieval scores compared to standard retrieval.
Despite that, \method consistently outperforms baselines across all training configurations and even surpasses the CLIP and SigLIP models trained on billions of samples. Interestingly, \method trained on CC12M-recap achieves higher R@1 scores (38.7\%), in terms of T2I retrieval on IIW-FG, compared to SigLIP-10B (33.8\%).

On the 30M dataset, the performance of \method further improves to 41.7\%, outperforming DreamLIP by 4.2\% in T2I retrieval (R@1). Overall, \method achieves an increased between 3.4\% and 7.8\% in R@1 scores compared to DreamLIP.
These results demonstrate that \method learns to align images with detailed, fine-grained captions more effectively than the baselines.

\subsection{Long Zero-shot Image-Text Retrieval}
\label{sec:retrieval_long}
Image-text retrieval with long captions imposes a unique challenge for CLIP models.
Following LoTLIP~\cite{wu2024lotlip} and Long-CLIP~\cite{zhang2024long}, we evaluate \method on datasets with long captions, including DCI~\cite{Urbanek_2024_CVPR}, 1k (SV-1k) and 10k (SV-10k) subsets of ShareGPT-4V~\cite{chen2023sharegpt4v}, and Urban-1k~\cite{zhang2024long}, with the results presented in~\cref{tab:long_retrieval_small}.

Unlike previous methods specifically designed for long-caption retrieval with extended token limits and larger text encoders, \method employs the standard CLIP text encoder with a 77-token limit. The former SOTA, LoTLIP~\cite{wu2024lotlip}, was trained on a 100M-scale re-captioned dataset, while Long-CLIP~\cite{zhang2024long} fine-tunes a 400M-scale CLIP model with an additional 1M images with long captions. Although \method is trained on a smaller training set of 30M samples, it still outperforms these methods on SV-1k, SV-10k, and Urban-1k. Most notably on T2I, \method obtained improvements of 10.4\%, 5.7\%, and 8.2\% in terms of R@1 over the previous SOTA. Remarkably, \method trained on 15M samples already surpasses all previous methods on 3 out of 4 datasets.

This significant performance gain can be explained as follows: 1) text-conditioned attention pooling can adapt to the rich semantics in long texts to extract all relevant visual information. 2) By sampling diverse captions
the model becomes aligned with the distribution of long captions.

\subsection{Zero-shot Semantic Segmentation}
\label{sec:segmentation}

For VLMs, zero-shot semantic segmentation
involves measuring the similarity \(\left\{ \langle \mathbf{v}^{\text{loc}}_{i}, \mathbf{t}^{\text{g}}_{j} \rangle \mid j \in \{1, 2, \dots, M\} \right\}\) for \( M \) different class names. Recent works~\cite{wang2023sclip, lan2024clearclip, dong2023maskclip} provide a framework to map these similarities to semantic segmentation outputs.
To examine the raw alignment of local image tokens $\mathbf{v}^{\text{loc}}$ with the corresponding input texts, we perform semantic segmentation following \cite{wang2023sclip} without post-processing or segmentation-specific modifications.

As shown in \cref{tab:ss}, \method trained on all subsets of data, consistently outperforms CLIP-based methods trained on significantly larger datasets, achieving an improvement of 10.1\% - 25.8\% mIOU increase across all datasets (14.4\% on average). As illustrated in \cref{fig:teaser_small}, DreamLIP’s \(\mathbf{v}^{\text{loc}}\) image token embeddings show weak correspondence to the input text, which we conclude is the result of their choice of negatives as discussed in \cref{subsec:textcon_pool}.
In contrast, \method, optimized with \(\mathcal{L}^{\text{tcs}}\), effectively aligns \(\mathbf{v}^{\text{loc}}\) with varying text prompts, demonstrating strong localization capabilities.

\subsection{Zero-shot Image Classification}
\label{sec:classification}
Following \cite{zheng2025dreamlip,fan2024improving}, we evaluate the zero-shot classification performance of \method and baseline methods on ImageNet~\cite{deng2009imagenet} and 10 additional datasets, as shown in \cref{tab:zs_cla}.
In retrieval tasks, text-augmented methods outperform VLMs trained on billions of images. However, in image classification they lag behind by around 20\%. This demonstrates that scaling up the number of images remains a key factor in improving VLM's classification performance.

When trained on CC3M-recap, \method achieves a 4.1\% higher average performance than DreamLIP and other baselines. This shows that \method, although optimized to generate fine-grained visual representations, could still efficiently gain global-level visual understanding performance when images are relatively scarce. However, when scaled up to 30M samples, \method, while still outperforming CLIP and SigLIP by 4\%, is on par with DreamLIP (-0.2\% on avg.). This shows that these methods trained on synthetic captions converge similarly, further suggesting the importance of scaling up images for the classification task. Therefore,
we hypothesize that scaling \method to larger datasets would extend the concept vocabulary and image coverage, closing the gap to large-scale models on zero-shot classification.

\subsection{Attention Maps Visualization}
\label{sec:qualitative}
\begin{figure}[t]
  \centering
   \includegraphics[width=\linewidth]{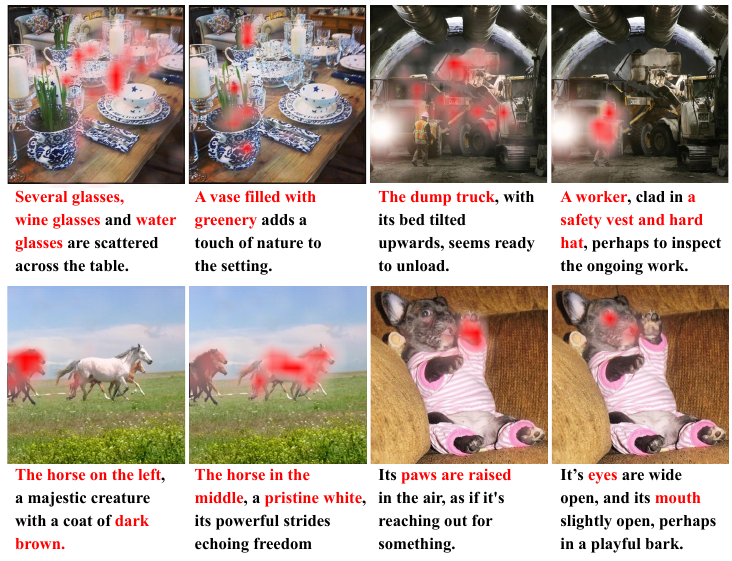}

   \caption{Visualization of attention maps in the attention pooling layer $f_{\text{AttnPool}}(.)$. Regions of high attention are highlighted in red.}
   \label{fig:attn_weights_vis}
\end{figure}

For a given image with two different local captions, we visualize the attention maps of \( f_{\text{AttnPool}}(.) \), i.e., which image tokens are pooled together, in \cref{fig:attn_weights_vis}. As illustrated by the ``truck'' and ``worker'' example, \method can locate both large and small objects in an image. The horses example shows that \method is able to differentiate objects by their individual properties such as color and location.
Notably, \method is also precise in identifying individual parts of an object, exemplified by the eyes, mouth, and paw of the dog, where the focus lies on the one that is raised.
These results show \method's strong sensitivity to semantic details.

\subsection{Ablation Study}
\label{sec:ablation_study}
\myparagraph{Model Components.}
In \cref{tab:main_ablation}, we analyze the components of \method: text conditioning (TC), global loss (GL), multiple captions per image (MC), and diverse caption sampling instead of single sentences (DS). 
\(\mathcal{L}^{\text{tcs}}\) and \(\mathcal{L}^{\text{mps}}\) correspond to TC+MC and GL+MC respectively.

SigLIP is equivalent to only using GL (1). Replacing GL with TC (2) leads to performance improvements across all metrics, achieving a 3.7\%/5.5\% increase in R@1 for COCO retrieval and a 33.8\% boost in VOC20 segmentation demonstrating its contribution to the fine-grained alignment.
Adding MC improves performance in both scenarios (3 and 4). Our diverse sampling (DS) is another significant improvement, especially in segmentation and long-retrieval performance on the Urban-1K which gains over 20\% (5 and 6).
\method, combining all components, achieves the best performance in all but long-retrieval (7).
In summary, \(\mathcal{L}^{\text{tcs}}\) is foundational to our method’s performance, sampling diverse captions provides a substantial boost in long retrieval tasks, while combining \(\mathcal{L}^{\text{mps}}\) delivers additional gains, particularly for global-level tasks.

\begin{table}[t]
	\centering\scriptsize
	\SetTblrInner{rowsep=1.2pt}
	\SetTblrInner{colsep=1.5pt}
	\resizebox{\columnwidth}{!}{
		\begin{tblr}{
			cells={halign=c,valign=m},   %
			hline{1,10}={1.0pt},       %
					hline{3}={},       %
					hline{2}={2-5}{leftpos = -1, rightpos = -1, endpos},
					hline{2}={6-7}{leftpos = -1, rightpos = -1, endpos},
					hline{2}={8-9}{leftpos = -1, rightpos = -1, endpos},
					hline{2}={10-11}{leftpos = -1, rightpos = -1, endpos},
					hline{2}={12-12}{leftpos = -1, rightpos = -1, endpos},
					hline{2}={13-13}{leftpos = -1, rightpos = -1, endpos},
					cell{1}{2}={c=4}{},
					cell{1}{6}={c=2}{},
					cell{1}{8}={c=2}{},
					cell{1}{10}={c=2}{},
				}
			  & Method      &             &             &             & COCO &          & DOCCI &          & Urban-1K &          & VOC20    & ImageNet \\
			& \textbf{GL} & \textbf{TC} & \textbf{MC} & \textbf{DS} & T2I          & I2T    & T2I           & I2T    & T2I              & I2T    & mIOU     & Top-1    \\
			1 & \checkmark  &             &             &             & 28.3           & 40.1     & 10.4            & 24.9     & 42.8               & 40.5     & 3.1      & 25.4     \\
			2 &            & \checkmark  &             &             & 32.0           & 45.6     & 12.7            & 30.9     & 44.4               & 42.6     & 36.9     & 28.1     \\
			3 & \checkmark  &             & \checkmark  &             & 32.9           & 44.6     & 13.3            & 31.0     & 47.9               & 46.5     & 1.7      & 27.9     \\
			4 &            & \checkmark  & \checkmark  &             & 34.8           & 47.1     & 14.1            & 30.2     & 46.6               & 40.9     & 34.1     & 29.4     \\
                5 &    \checkmark        &  & \checkmark  &     \checkmark        &     35.0     &  49.1   &     13.0       &  33.1    &      \bf 70.7       & \bf 64.6   &  7.2  &  32.0   \\
			6 &            & \checkmark  & \checkmark  & \checkmark  & 36.2           & 50.0     & 13.8            & 34.6     & 68.3               & 63.1     & 46.5     & 31.5     \\
			7 & \checkmark  & \checkmark  & \checkmark  & \checkmark  & \bf 37.7       & \bf 51.6 & \bf 15.1        & \bf 35.7 & 69.5           & 63.5 & \bf 59.7 & \bf 33.8 \\
		\end{tblr}}
	\caption{Ablation study on different components of \method on the CC3M-recap dataset. \textbf{GL}: Global Loss, \textbf{TC}: Text Conditioning, \textbf{MC}: Multiple Captions, \textbf{DS}: Diverse Sampling.
    }
	\label{tab:main_ablation}
\end{table}

\myparagraph{Additional Ablations.}
In supplementary Sec.~\ref{sec:supp_ablation}, we provide additional ablations to support important choices.
We pre-trained \method on the original CC3M dataset and on the PixelProse~\cite{singla2024pixels} dataset with synthetic captions generated by Gemini-Pro~\cite{reid2024gemini}, showing that our method is not restricted to long captions and adaptable to a variety of data distributions.
By varying the sampling strategy of the diverse captions, we find that it is crucial across tasks to sample both short and long captions instead of only a fixed length.
By testing a different number of multiple captions $K$ ranging from 2 to 10,
we observe that performance converges at around 8 captions. 
Finally, we ablate the maximal number of sampled sentences $S$
and observe that merging 3 sentences achieved the most balanced results.

    \section{Conclusion and Limitations}

We introduce \method, a VLM that learns \fullname by conditioning on the semantics in dense local captions. Trained on 30M recaptioned images, \method outperforms baselines trained on billions of images across standard, fine-grained, and long-form image-text retrieval tasks. The significant improvements in zero-shot segmentation compared to the baselines as well as the qualitative results corroborate that \method learns a fine-grained alignment between text and image at the token-level.

While \method matches baselines trained on the same number of images in zero-shot classification, it still falls behind CLIP models trained on significantly larger datasets. This suggests that, although leveraging detailed synthetic captions enhances fine-grained image understanding, it does not replace the image coverage and conceptual richness of larger datasets for global-level tasks. To tackle this limitation, a natural future direction involves scaling the synthetic recaptioning to large-scale datasets and training variants of \method with higher parameter count.
\\~\\

    \myparagraph{Acknowledgments}
This work was partially funded by the ERC (853489 - DEXIM) and the Alfried Krupp von Bohlen und Halbach Foundation, which we thank for their generous support. The authors gratefully acknowledge the Gauss Centre for Supercomputing e.V. (\url{www.gauss-centre.eu}) for funding this project by providing computing time on the GCS Supercomputer JUWELS~\cite{JUWELS} at Jülich Supercomputing Centre (JSC).

    {\small
     \bibliographystyle{ieeenat_fullname}
     \bibliography{main}
     }
    \clearpage
\maketitlesupplementary
\appendix

In this supplementary file, we illustrate more qualitative results in Sec.~\ref{sec:qualitative_supp}, describe the datasets in Sec.~\ref{sec:example_sampled_captions}, and present an extensive analysis of the impact of the negative pairs on the \method performance in Sec.~\ref{sec:supp_negatives}. We further present additional ablation experiments in Sec.~\ref{sec:supp_ablation}, and the implementation details in Sec.~\ref{sec:implementation_details}.

\section{Qualitative Results}
\label{sec:qualitative_supp}

\begin{figure}[t]
    \centering
    \includegraphics[width=\linewidth]{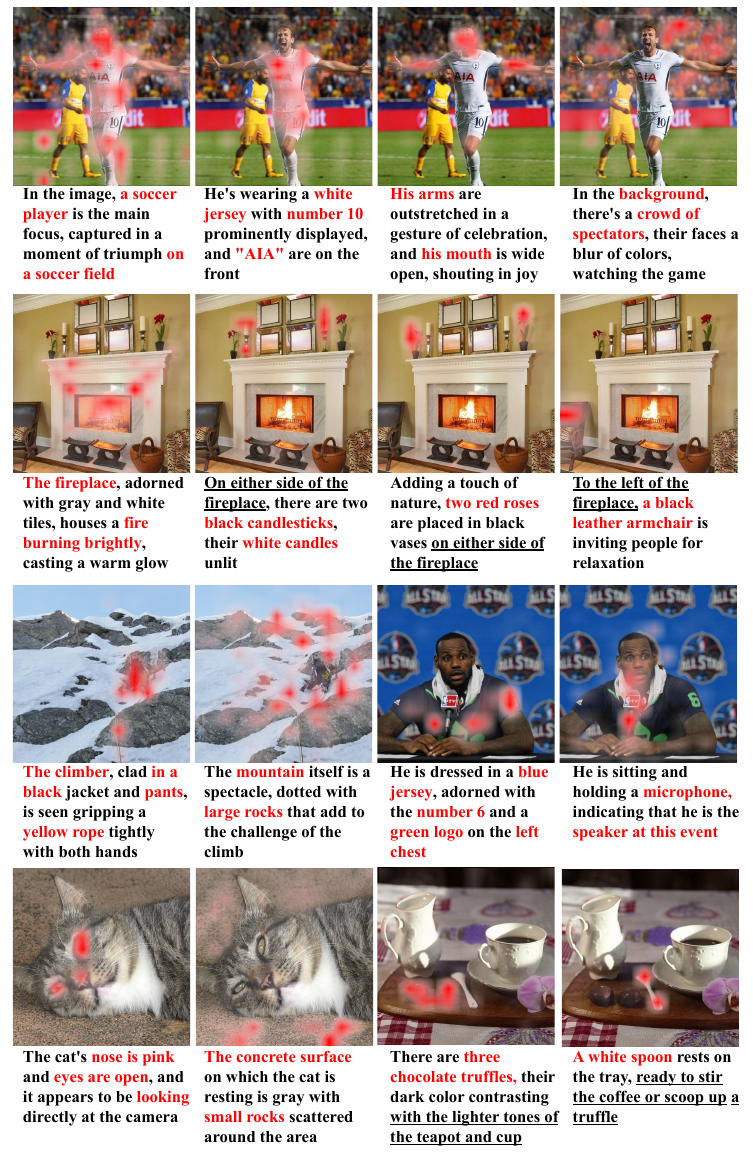}
\caption{Visualization of the attention maps w.r.t. fine-grained captions. In the images, regions with high attention scores are marked in red; in the captions, objects representing the main semantics of the sentences are marked in red, while objects with less semantic significance are \underline{underlined}.
}
    \label{fig:appendix_attn_map_short}
\end{figure}

\begin{figure}[t]
    \centering
    \includegraphics[width=\linewidth]{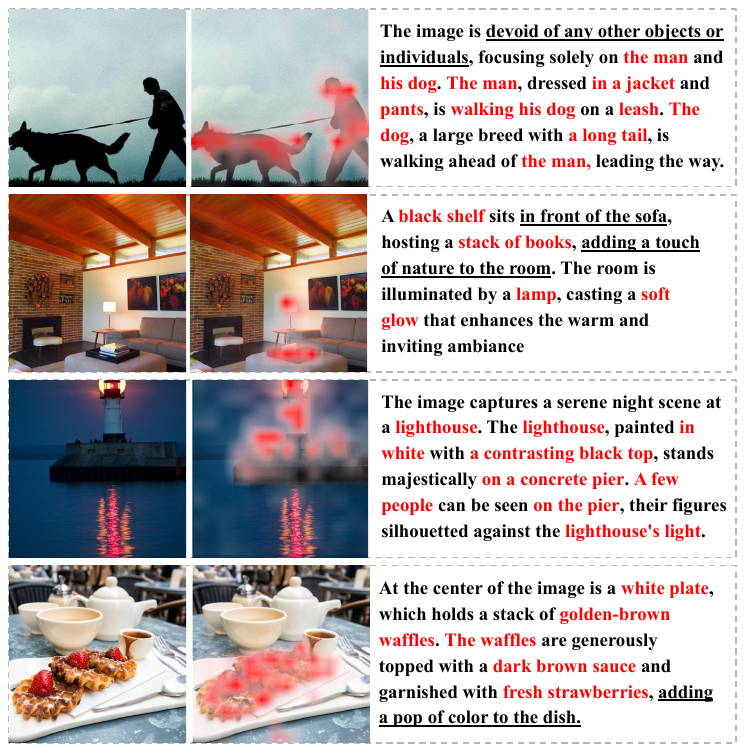}
\caption{Visualization of the attention maps w.r.t. fine-grained long captions. In the images, regions with high attention scores are marked in red; in the cap
tions, objects representing the main semantics of sentences are marked in red, while objects with less semantic significance are \underline{underlined}.
}
    \label{fig:appendix_attn_map_long}
\end{figure}

\noindent
\myparagraph{Attention Maps Visualization.}
We provide a comprehensive visualization of attention maps of $f_{\text{AttnPool}}(.)$ in \cref{fig:appendix_attn_map_short} and \cref{fig:appendix_attn_map_long}. We follow DINO~\cite{caron2021emerging} to aggregate attention maps from multiple heads. We empirically found that heads 1,4,6,8 mainly focus on foreground objects and aggregate these attention maps to form the visualization. In \cref{fig:appendix_attn_map_short}, we show that the attention maps focus on different parts of an image w.r.t. the local captions. Interestingly, in the ``fireplace'' example (second row), the attention correctly localizes the ``white candle'' (second row, second column), which is exactly what the caption describes, although ``fireplace'' also appears in the sentence. This demonstrates that \method is able to locate an object based on the main semantics of a prompt, instead of simply matching ``a bag of words''. 

In \cref{fig:appendix_attn_map_long}, we visualize the attention maps w.r.t. long captions. When multiple objects appear in a long caption, \method  is able to locate them at the same time. Notably, in the ``room'' example (second row), \method ignores descriptions like ``adding a touch of nature to the room'' and solely focus on the main semantics: ``black shelf'', ``books'' and ``lamp''. This might reveal one possible future application of \method, understanding the main semantics in complex prompts and grounding the main objects in the image. 

\begin{figure}[t]
  \centering
   \includegraphics[width=\linewidth]{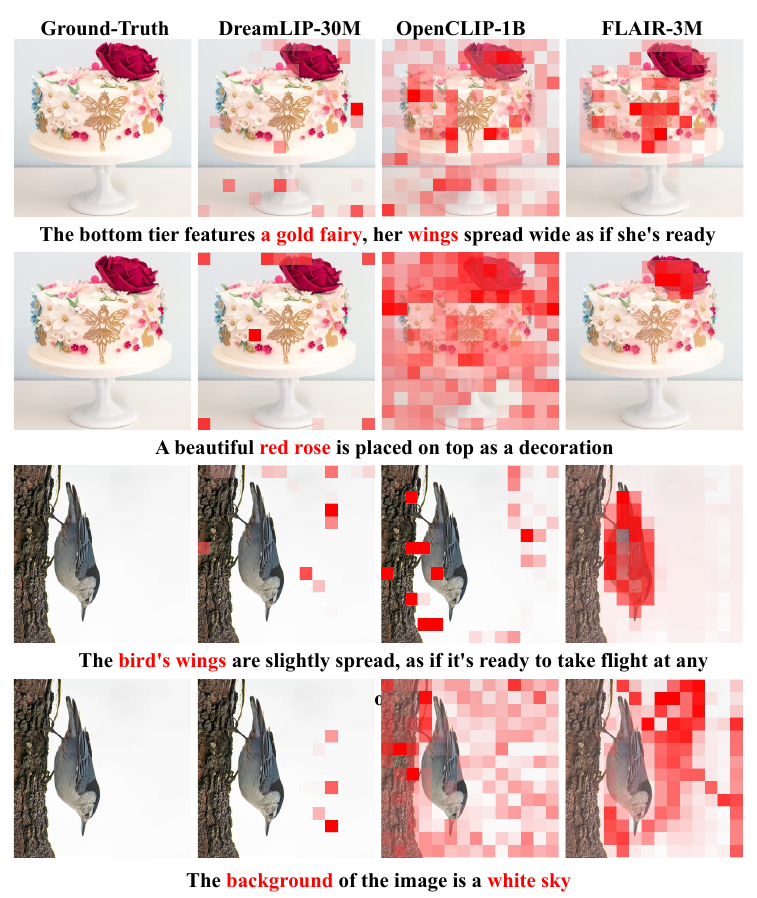}

   \caption{Visualization of the similarity scores between local image tokens and different text queries.
   While previous works~\cite{zheng2025dreamlip,cherti2023reproducible} lack fine-grained alignment,
   \method matches text and image semantics at the token level.
   }
   \label{fig:tokenwise_sim_appendix}
\end{figure}

\noindent
\myparagraph{Token-to-Text Similarity.}
We also visualize the similarity between local image tokens and text prompts in Fig. 1 of the main paper. This similarity between the local image tokens and the text prompts could reflect the model's localization capability, which is closely related to the segmentation task. We provide extra visualizations in \cref{fig:tokenwise_sim_appendix}. We use \method pre-trained on the CC3M-recap dataset to compare with DreamLIP~\cite{zheng2025dreamlip} trained on Merged30M dataset and OpenCLIP trained on DataComp-XL~\cite{gadre2024datacomp}. As illustrated, compared to OpenCLIP~\cite{cherti2023reproducible} that tends to make over-predictions, \method is able to accurately localize the tokens w.r.t. the text prompts, especially on fine-grained details such as ``flower on the cake'' and ``bird on the branch''. This further validates that the fine-grained representations learned by \method are indeed sensitive to the text semantics.

\begin{figure}[t]
  \centering
   \includegraphics[width=\linewidth]{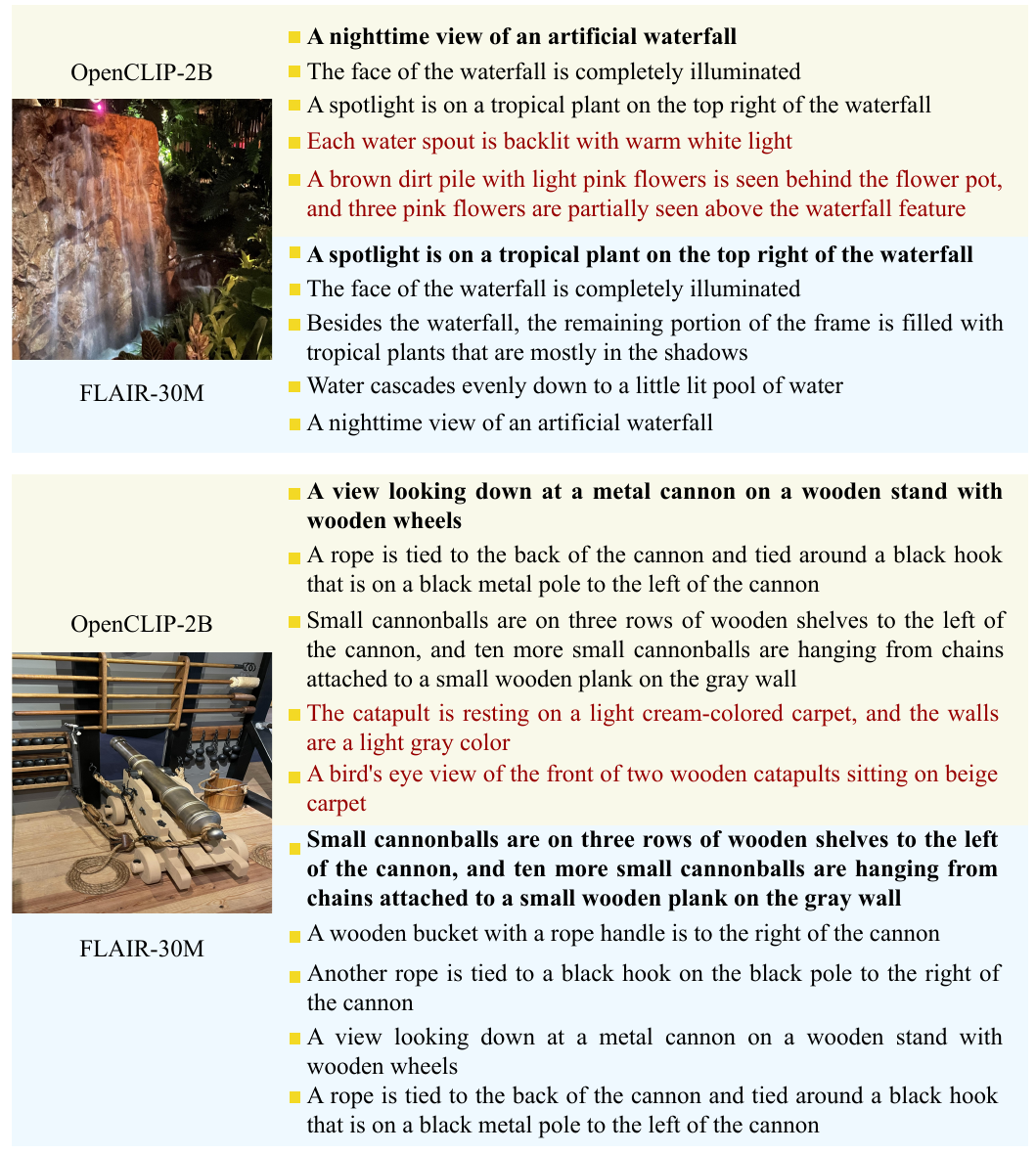}

   \caption{Visualization of image-to-text retrieval samples on the DOCCI-FG~\cite{onoe2024docci} benchmark, comparing \method with OpenCLIP~\cite{cherti2023reproducible}. For each image, the top-5 retrieved captions are displayed. The incorrect retrieved captions are marked in red. The top-1 retrieved captions are \textbf{bold}. 
   }
   \label{fig:i2t_retrieval_appendix}
\end{figure}

\noindent
\myparagraph{Retrieval Visualization.}
For the fine-grained image-text retrieval task on the DOCCI~\cite{onoe2024docci} benchmark, we visualize the top-5 retrieved captions for a given image, highlighting incorrect captions in red. We compare \method with OpenCLIP~\cite{cherti2023reproducible} trained on 2B samples in \cref{fig:i2t_retrieval_appendix}.  From top to bottom, the similarity scores decrease. Interestingly, compared to OpenCLIP~\cite{cherti2023reproducible}, \method tends to retrieve ``local'' captions first. For example, the top-1 retrieved caption for \method is only describing the ``spotlight'', while OpenCLIP retrieves ``a nighttime view of an artificial waterfall'', which can be considered a global description for this image. The incorrectly retrieved captions of OpenCLIP contain relevant keywords like ``waterfall'', while \method retrieves the captions correctly based on a more detailed understanding of the image semantics. 

\section{Dataset Details}
\label{sec:example_sampled_captions}

\begin{figure*}[t]
  \centering
   \includegraphics[width=\linewidth]{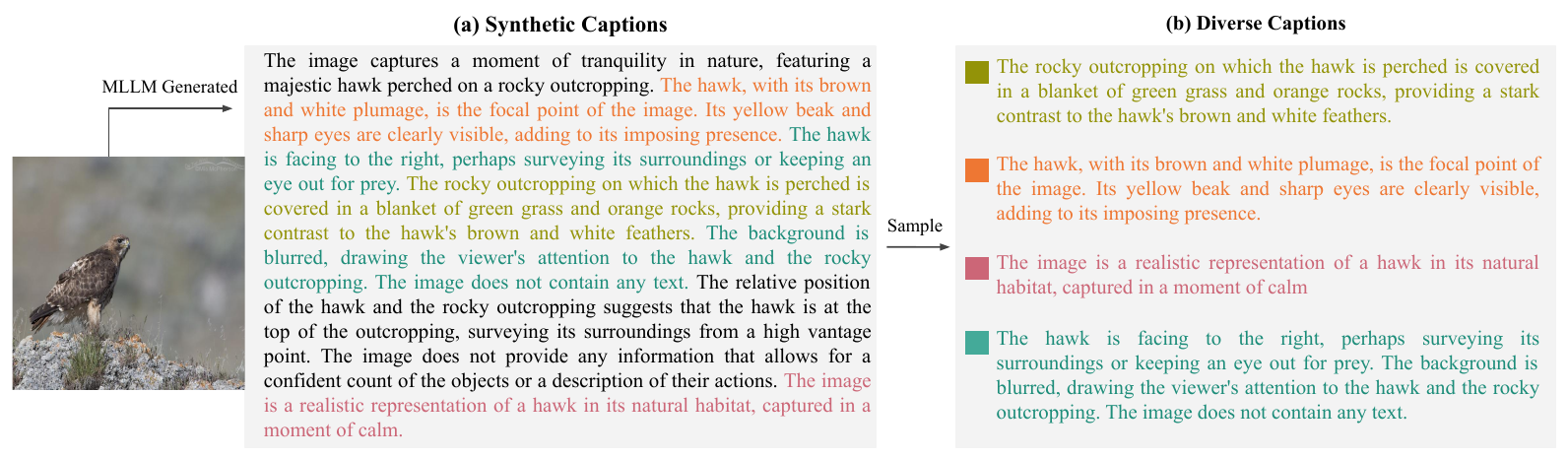}

   \caption{Examples of our diverse captions. Image and captions are taken from CC3M-recap~\cite{zheng2025dreamlip}. Given the synthetic long captions generated by an MLLM, here we sample $K=4$ sub-captions where each sub-caption consists of $s \in \{1,2,3\}$ sentences. In our main experiments, we use $K=8$.
   }
   \label{fig:example_training_data}
\end{figure*}

\myparagraph{Pre-training Data.} \method is pre-trained on CC3M-recap, CC12M-recap, YFCC15M-recap and Merged-15M~\cite{zheng2025dreamlip}, where each image is equipped with long synthetic captions generated by various MLLMs. \cref{fig:example_training_data} shows an example of the original long captions produced by DreamLIP~\cite{zheng2025dreamlip} together with our diverse sampled captions. We take the whole paragraph of the long synthetic caption and split it into sentences. Our $K$ diverse captions are sampled from these sentences, and each caption can contain $s \in \{ 1,..., S\}$ merged sentences. In our experiments, we set $S=3$ and $K=8$. We detail this choice in Sec.~\ref{sec:num_sampled_captions} and Sec.~\ref{sec:random_merged_cap}. 

\begin{figure}[t]
    \centering
    \includegraphics[width=\linewidth]{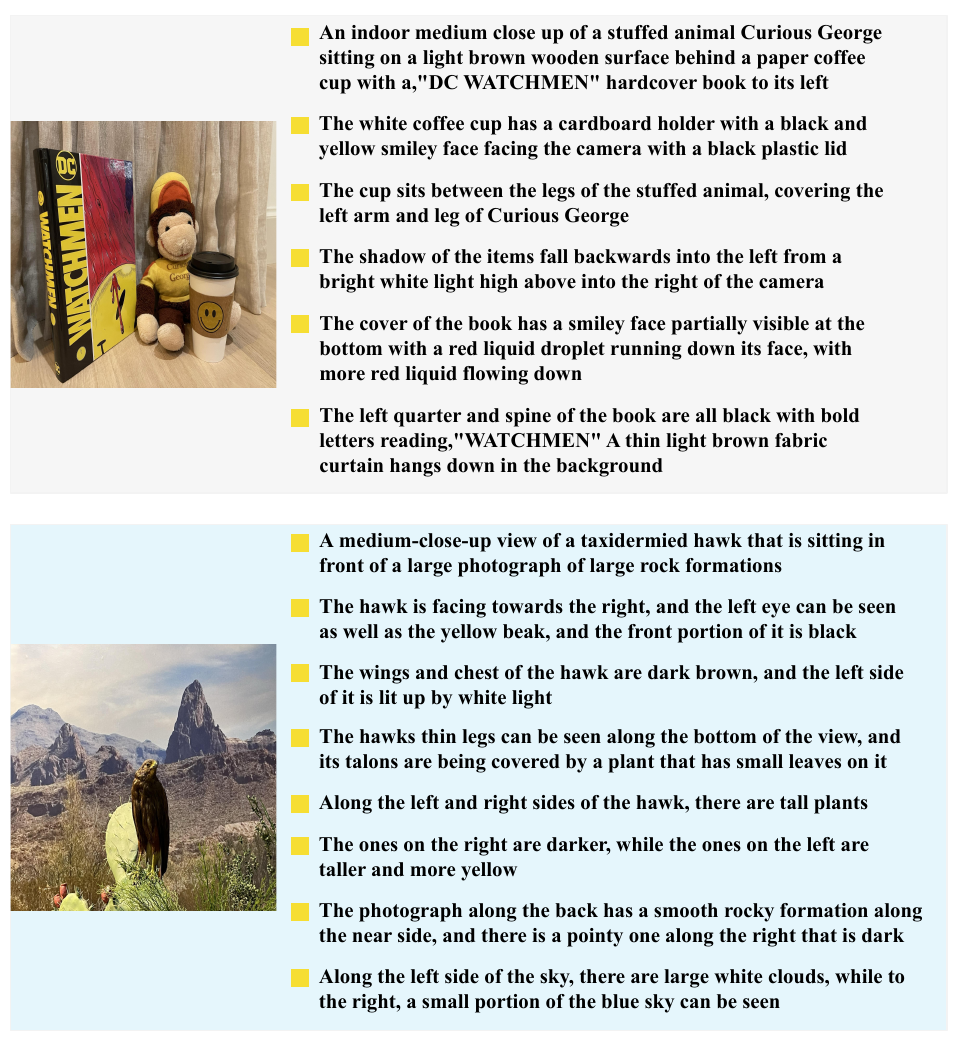}
\caption{Dataset samples from DOCCI-FG~\cite{onoe2024docci} for the fine-grained retrieval task. For each image, we split the long caption into individual sentences each serving as a positive image-text pair for the benchmark. 
}
    \label{fig:docci_example}
\end{figure}

\myparagraph{Fine-grained Retrieval Data.}
In order to create the new fine-grained retrieval task, we split the original long captions from DOCCI~\cite{onoe2024docci} and IIW~\cite{garg2024imageinwords} into separate sentences. Each sentence can either describe the image globally or describe the fine-grained details of an image. These captions, together with the original images, form our DOCCI-FG and IIW-FG retrieval benchmarks. We provide a visualization of DOCCI-FG containing two images with all the corresponding paired captions in \cref{fig:docci_example}. As illustrated in \cref{fig:docci_example}, the split captions are likely to describe one local part of an image, such as ``The wings and chest of the hawk are dark brown, and the left side of it is lit up by white light''. We provide detailed statistics on the number of images, captions, and the average number of tokens per caption for standard, fine-grained, and long retrieval benchmarks in \cref{tab:retrieval_data_statistics}. DOCCI-FG and IIW-FG contain an average of 7.1 and 10.1 captions per image, respectively, with each caption comprising approximately 18.76 and 22.56 tokens.

\begin{figure}[t]
    \centering
    \includegraphics[width=\linewidth]{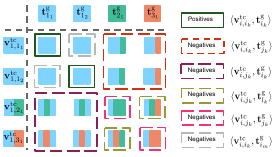}
\caption{Illustration on all possible positive and negative pairs for \method.
}
    \label{fig:negatives_explanation}
\end{figure}

\begin{table}[t]
    \centering
    \scriptsize
    \SetTblrInner{rowsep=1.2pt}
	\SetTblrInner{colsep=2pt}
    \resizebox{\linewidth}{!}{
    \begin{tblr}{
        cells={halign=c,valign=m},   
        hline{1,13}={1.0pt},         
         hline{2,3,5, 6, 8, 9}={},
        cell{2,5,8}{1}={c=5}{},
    }
         Dataset & \#Images & \#Captions  & \#Captions per Image & \#Tokens per Caption  \\
         Standard Text-image Retrieval Dataset \\
         MSCOCO~\cite{lin2014microsoft} & 5,000& 25,000 & 5.0 &11.77 \\
         Flickr30K~\cite{plummer2015flickr30k} & 1,000& 5,000 & 5.0 & 14.03 \\
         Fine-grained Text-image Retrieval Dataset \\
         DOCCI-FG~\cite{onoe2024docci} & 5,000 & 35,533 & 7.1 &  18.76\\
         IIW-FG~\cite{garg2024imageinwords} & 612 & 6204 & 10.1 & 22.56 \\
         Long Text-image Retrieval Dataset \\
         DCI~\cite{Urbanek_2024_CVPR} & 7,805& 7,805  & 1.0  & 172.73 \\
         IIW~\cite{garg2024imageinwords} & 612& 612 & 1.0 & 239.73\\
         SV-1k~\cite{chen2023sharegpt4v} & 1,000& 1,000 &  1.0  & 173.24 \\
         SV-10k~\cite{chen2023sharegpt4v} &10,000&10,000&  1.0  & 173.66\\         
        \end{tblr}
        }
    \caption{Dataset details of the standard, fine-grained and long image-text retrieval task. SV-1K and SV-10K denote the 1K and 10K subset from the ShareGPT4V~\cite{chen2023sharegpt4v} dataset. Values of long text-image retrieval are directly obtained from \cite{wu2024lotlip}, since we follow their evaluation setting.}\label{tab:retrieval_data_statistics}
\end{table}

\section{Extended Analysis of Negatives}\label{sec:supp_negatives}

As discussed in the methodology section of the main paper, \method produces a unique image representation for each image-text pair using the text-conditioned attention pooling. Specifically, the text-conditioned embedding \(\mathbf{v}^{\text{tc}}\) is jointly conditioned by the local image tokens \(\mathbf{v}^{\text{loc}}\) and global text tokens \(\mathbf{t}^{\text{g}}\):
\[
\mathbf{v}^{\text{tc}} = f_{\text{AttnPool}}(\mathbf{v}^{\text{loc}}, \mathbf{t}^{\text{g}})
\]
When considering the global text token \(\mathbf{t}^{\text{g}}\), which forms both positive and negative pairs in \(\mathcal{L}^{\text{tcs}}\), one positive pair (\(\langle \mathbf{v}^{\text{tc}}_{i,i_k}, \mathbf{t}^{\text{g}}_{i_k} \rangle\)) and five types of negative pairs can be identified. As visually depicted in \cref{fig:negatives_explanation}, these negatives are:
\[
\langle \mathbf{v}^{\text{tc}}_{i,j_k}, \mathbf{t}^{\text{g}}_{j_k} \rangle, 
\langle \mathbf{v}^{\text{tc}}_{i,j_k}, \mathbf{t}^{\text{g}}_{i_k} \rangle,
\langle \mathbf{v}^{\text{tc}}_{i,i_k}, \mathbf{t}^{\text{g}}_{j_k} \rangle,
\langle \mathbf{v}^{\text{tc}}_{i,j_k}, \mathbf{t}^{\text{g}}_{l_k} \rangle, 
\langle \mathbf{v}^{\text{tc}}_{i,i_k}, \mathbf{t}^{\text{g}}_{i_m} \rangle
\]
The notation \(\{i, j, l\}\) indicates that this pair is constructed from the \(\{\text{Image, Text Condition, Text}\}\), which stems from the \(\{i\text{-th}, j\text{-th}, l\text{-th}\}\) image separately, while \(k\) represents the \(k\text{-th}\) caption for image \(i\). The pair \(\langle \mathbf{v}^{\text{tc}}_{i,i_k}, \mathbf{t}^{\text{g}}_{i_m} \rangle\) is unique, as it arises from the \(k\text{-th}\) and \(m\text{-th}\) captions of the same image.

\noindent
\myparagraph{Empirical Comparison.} 
By introducing text-conditioned attention pooling for multi-caption settings, \method considers one positive and up to five distinct negative pairings. Modeling all five negatives simultaneously causes significant computational overhead. Thus, we investigate the importance of each negative type. To study their effects, we conducted a comprehensive ablation experiment (\cref{tab:negative_type_performance}). For each setup, we trained \method with one positive and only one negative pairing at a time, using a batch size of 1,024. All models were trained on the CC3M-recap~\cite{zheng2025dreamlip} dataset for 10 epochs.

To evaluate training dynamics, we analyzed the training loss (\(\mathcal{L}_{\text{train}}\)) and validation performance using the MSCOCO retrieval task. Key findings include:
1. The negative \(\langle \mathbf{v}^{\text{tc}}_{i,j_k}, \mathbf{t}^{\text{g}}_{l_k} \rangle\) suffers from high \(\mathcal{L}_{\text{train}}\) and poor validation performance. As this negative spans across three different source images, it likely introduces noise rather than aiding learning.
2. The negatives \(\langle \mathbf{v}^{\text{tc}}_{i,j_k}, \mathbf{t}^{\text{g}}_{i_k} \rangle\) and \(\langle \mathbf{v}^{\text{tc}}_{i,i_k}, \mathbf{t}^{\text{g}}_{i_m} \rangle\) converge quickly during training, but their \(\mathcal{L}_{\text{train}}\) swiftly drops to nearly zero. Their evaluation on MSCOCO reveals poor performance, suggesting the existence of shortcuts. For \(\langle \mathbf{v}^{\text{tc}}_{i,j_k}, \mathbf{t}^{\text{g}}_{i_k} \rangle\), the model likely ignores image information and relies solely on text conditioning, thus failing in evaluation, when image information is vital.
3. The negative \(\langle \mathbf{v}^{\text{tc}}_{i,i_k}, \mathbf{t}^{\text{g}}_{j_k} \rangle\) converges to a reasonable \(\mathcal{L}_{\text{train}}\), but its performance (2.4\% R@1 on T2I) indicates limited learning benefit.
4. The negative \(\langle \mathbf{v}^{\text{tc}}_{i,j_k}, \mathbf{t}^{\text{g}}_{j_k} \rangle\), currently used in \method, reaches the best retrieval results, demonstrating its effectiveness.

\begin{table}[t!]
     \centering\scriptsize
     \SetTblrInner{rowsep=1.2pt}
     \SetTblrInner{colsep=2pt}
     \resizebox{.8\linewidth}{!}{
     \begin{tblr}{
         cells={halign=c,valign=m},
         column{1}={halign=l},
         hline{1,7}={1-7}{1pt},
         hline{2}={1-7}{}
     }
         Neg.  & $\mathcal{L_{\text{train}}}$ & T2I@1 & T2I@5 & I2T@1 & I2T@5  \\
         $\langle \mathbf{v}^{\text{tc}}_{i,j_k}, \mathbf{t}^{\text{g}}_{l_k} \rangle$  & 5.8 & 0.0 & 0.1 &0.0 &0.1\\
         $\langle \mathbf{v}^{\text{tc}}_{i,i_k}, \mathbf{t}^{\text{g}}_{i_m} \rangle$  &  0.0  &  0.0  & 0.1 & 0.0 & 0.0\\
         $\langle \mathbf{v}^{\text{tc}}_{i,j_k}, \mathbf{t}^{\text{g}}_{i_k} \rangle$ &  0.0  &  0.0  & 0.1 & 0.0 & 0.0\\
          $\langle \mathbf{v}^{\text{tc}}_{i,i_k}, \mathbf{t}^{\text{g}}_{j_k} \rangle$&   1.53 &  2.4 & 7.8 & 0.3 & 1.2   \\
           $\langle \mathbf{v}^{\text{tc}}_{i,j_k}, \mathbf{t}^{\text{g}}_{j_k} \rangle$  &    0.68  &  \textbf{24.5} & \textbf{49.1} & \textbf{36.4 }& \textbf{62.7}\\
     \end{tblr}}
     \caption{Retrieval performance if \method on the MSCOCO~\cite{lin2014microsoft} validation set when trained with different negative types on the CC3M-recap~\cite{zheng2025dreamlip} dataset for 10 epochs. All models use ViT-B/16 as vision encoder. The best retrieval results are \textbf{bold}.}
     \label{tab:negative_type_performance}
\end{table}

\section{Additional Ablation Experiments}
\label{sec:supp_ablation}
Aside from the main ablation study on the components of \method described in the main paper, we conduct additional experiments to validate specific design choices. These include pre-training \method on different data sources (Sec.~\ref{sec:different_sources}), comparing the diverse sampling strategy with a fixed merging strategy (Sec.~\ref{sec:sampling_strategy}), ablating the maximum number of sampled sentences $S$ (Sec.~\ref{sec:random_merged_cap}), and examining how the number of sampled captions $K$ affects the performance (Sec.~\ref{sec:num_sampled_captions}).

\subsection{Pre-training on Different Data Sources}
\label{sec:different_sources}

To demonstrate that our model is not limited to data curated by DreamLIP~\cite{zheng2025dreamlip}, we also pre-train \method on the original CC3M~\cite{sharma2018conceptual} (CC3M-orig) and PixelProse~\cite{singla2024pixels}. For CC3M-orig and PixelProse, we use the same pre-training configurations as CC3M-recap and CC12M-recap, respectively. Detailed configurations are available in Sec.~\ref{sec:implementation_details}. CC3M-orig contains one conceptual caption per image, while PixelProse re-captioned 15M images from CC12M~\cite{changpinyo2021cc12m}, RedCaps~\cite{desai2021redcaps}, and CommonPool~\cite{gadre2024datacomp} using Gemini-Pro~\cite{reid2024gemini}. Unlike DreamLIP, which uses three MLLMs for re-captioning, PixelProse employs a single MLLM, resulting in shorter captions.

We evaluate \method on the standard retrieval task and compare its performance to CLIP~\cite{radford2021learning} trained on the same datasets. The results are summarized in \cref{tab:retrieval_different_sources}.

As shown in \cref{tab:retrieval_different_sources}, even when pre-trained on CC3M-orig, where \method cannot leverage additional augmented captions, it still achieves a 2\% improvement over CLIP in terms of R@1 on the MSCOCO dataset~\cite{lin2014microsoft}. This demonstrates that \method is capable of effectively enhancing the retrieval performance even on datasets with only global captions. Furthermore, when pre-trained on PixelProse, \method achieves an 8\% improvement in both text-to-image (T2I) and image-to-text (I2T) retrieval tasks on MSCOCO. These results indicate that \method is versatile and can be applied to datasets where images are captioned by a different MLLM, while maintaining significant performance gains.

\begin{table}[t!]
    \centering\scriptsize
    \SetTblrInner{rowsep=1.2pt}
    \SetTblrInner{colsep=2pt}
    \resizebox{\linewidth}{!}{
    \begin{tblr}{
        cells={halign=c,valign=m},
        column{1}={halign=l},
        hline{1,8}={1-14}{1pt},
        hline{4}={1-14}{},
        cell{1}{1}={r=3}{},
        cell{1}{2}={r=3}{},
        cell{1}{3}={c=4}{},
        cell{1}{7}={c=4}{},
        cell{2}{3}={c=2}{},
        cell{2}{5}={c=2}{},
    }

        Data           & Method & \SetCell[c=4]{c} MSCOCO & & & &  \SetCell[c=4]{c} Flickr30k & & & &\\
                       &        & \SetCell[c=2]{c} T2I & &  \SetCell[c=2]{c} I2T & &  \SetCell[c=2]{c} T2I & &  \SetCell[c=2]{c} I2T & & \\
                       &        & R@1            & R@5  & R@1    & R@5  & R@1             & R@5  & R@1    & R@5  \\
        CC3M-orig~\cite{sharma2018conceptual}      & CLIP~\cite{radford2021learning} & 4.75  & 14.53 & 5.90 & 17.56 & 9.19 & 23.61  & 12.13 & 29.68 \\
                       & \method  & \bf 6.45  & \bf 18.14  & \bf 8.00 & \bf 22.48 & \bf 12.70 & \bf 30.20  & \bf 17.55 & \bf 38.66  \\
        PixelProse~\cite{singla2024pixels}     & CLIP~\cite{radford2021learning} & 28.86 & 54.05 & 48.50 & 74.24 & 54.06 & 77.81 & 79.09 & 94.87 \\
                       & \method & \bf 36.08 & \bf 61.18 & \bf 56.56 & \bf 79.06  & \bf 64.87 & \bf 85.03  & \bf 86.69 & \bf 97.14  \\
    \end{tblr}}
    \caption{Standard zero-shot image-text retrieval on the validation splits of Flickr30k~\cite{plummer2015flickr30k} and MSCOCO~\cite{lin2014microsoft}. CLIP and \method are pre-trained on CC3M-orig and PixelProse under the same training configurations, using ViT-B/16 as the vision encoder.}
    \label{tab:retrieval_different_sources}
    \vspace{-10pt}
\end{table}

\begin{table}[t]
	\centering\scriptsize
	\SetTblrInner{rowsep=1.2pt}
	\SetTblrInner{colsep=2pt}
	\resizebox{\columnwidth}{!}{
		\begin{tblr}{
			cells={halign=c,valign=m},  
			hline{1,6}={1.0pt},      
					hline{3}={},      
					hline{2}={2-3}{leftpos = -1, rightpos = -1, endpos},
					hline{2}={4-5}{leftpos = -1, rightpos = -1, endpos},
					hline{2}={6-7}{leftpos = -1, rightpos = -1, endpos},
					hline{2}={8-8}{leftpos = -1, rightpos = -1, endpos},
					hline{2}={9-9}{leftpos = -1, rightpos = -1, endpos},
					cell{1}{2}={c=2}{},
					cell{1}{4}={c=2}{},
					cell{1}{6}={c=2}{},
				}
         & MSCOCO & & DOCCI & & Urban-1K & & VOC20 & ImageNet \\
        \bf Merging & T2I@1 & I2T@1 & T2I@1 & I2T@1 & T2I@1 & I2T@1 & mIOU & Top-1 \\
        No & 35.8 & 47.1 & 14.8 & 33.2 & 46.4 & 42.4 & 52.2 & 29.9 \\
        Always & 34.2 & 46.8 & 12.4 & 30.1 & \textbf{70.9} & \textbf{64.7} & 54.9 & 27.7 \\
        Random & \textbf{37.7} & \textbf{51.6} & \textbf{15.1} & \textbf{35.7} & 69.5 & 63.5 & \textbf{59.7} & \textbf{33.8} \\
		\end{tblr}}
    \caption{Ablation study on merging strategies for sampling captions. No: only sample 1 sentence as the sampled caption. Always: always merge 3 sampled sentences into one caption. Random: each caption is merged randomly from 1-3 sentences. We train \method on CC3M-recap with 8 captions per image.}
    \label{tab:mixed_merging_ablation}
\end{table}
\subsection{Sampling Strategy}
\label{sec:sampling_strategy}
When sampling diverse captions, we randomly merge $s \in \{1,\dots,S\}$ sentences in the original MLLM-generated long captions to form a single caption. To evaluate this strategy, we compare \method with three settings: randomly merging 1–3 sentences, always merging 3 sentences, and no merging. The results, presented in \cref{tab:mixed_merging_ablation}, show that always merging 3 sentences improves Urban-1K T2I R@1 and I2T R@1 by 1.4\% and 1.2\%, respectively. However, it decreases T2I R@1 and I2T R@1 on MSCOCO by 3.5\% and 5.2\%, indicating a bias towards long retrieval tasks at the expense of short retrieval performance.

Conversely, random merging outperforms the no-merging setting across all metrics, effectively balancing short and long retrieval tasks. Additionally, it enhances model performance by introducing diverse data augmentations through caption variations.

\begin{table}[t]
	\centering\scriptsize
	\SetTblrInner{rowsep=1.2pt}
	\SetTblrInner{colsep=2pt}
    \resizebox{\columnwidth}{!}{
	\begin{tblr}{
		cells={halign=c,valign=m},  
        hline{1,6}={1.0pt},
        hline{3}={},
        hline{2}={2-3}{leftpos = -1, rightpos = -1, endpos},
        hline{2}={4-5}{leftpos = -1, rightpos = -1, endpos},
        hline{2}={6-7}{leftpos = -1, rightpos = -1, endpos},
        hline{2}={8-8}{leftpos = -1, rightpos = -1, endpos},
        hline{2}={9-9}{leftpos = -1, rightpos = -1, endpos},
        cell{1}{2}={c=2}{},
        cell{1}{4}={c=2}{},
        cell{1}{6}={c=2}{},
        }
        & MSCOCO & & DOCCI & & Urban-1K & & VOC20 & ImageNet \\
        \bf S & T2I@1 & I2T@1 & T2I@1 & I2T@1 & T2I@1 & I2T@1 & mIOU & Top-1 \\
        2 & 37.1 & 50.7 & 14.2 & 35.2 & 68.5 & 62.8 & 59.0 & 32.0 \\
        3 & \textbf{37.7}& 51.6 & \textbf{15.1} & \textbf{35.7} & \textbf{69.5} & \textbf{63.5} & \textbf{59.7} & \textbf{33.8} \\
        4 & 37.5 & \textbf{52.0} & 14.6 & 35.2 & \textbf{69.5} & 63.2 & 57.4 & 33.0 \\
	\end{tblr}
    }
    \caption{Ablation on the maximum number of sentences ($S$) to be merged for create a new sub-caption. We trained \method with $S \in [2,4]$ on CC3M-recap dataset under the same training configuration. The best results are \textbf{bold}.}
    \label{tab:s_ablation}
\end{table}

\subsection{Number of Merged Sentences}
\label{sec:random_merged_cap}
In the diverse caption sampling strategy, each new caption is created by merging up to $S$ sentences. In \cref{tab:s_ablation}, we train \method with $S=2$, $S=3$, and $S=4$. Compared to $S=2$, $S=3$ yields consistent improvements across all downstream tasks. However, increasing to $S=4$ does not lead to further gains, likely because merging four sentences often exceeds the 77-token limit of the text encoder. Based on these findings, we set $S=3$ for our main experiments.

\subsection{Number of Sampled Sub-captions}
\label{sec:num_sampled_captions}
In \cref{tab:k_ablation}, we pre-train \method with a different number of sampled captions $K$ ranging from 2 to 10 on the CC3M-recap dataset. We also compared CLIP~\cite{radford2021learning} and SigLIP~\cite{zhai2023sigmoid} pre-trained on the same dataset. First, even when $K=2$, \method surpasses SigLIP by 8.1\% (T2I R@1) and 9.0\% (I2T R@1) on MSCOCO retrieval. Increasing to $K=8$ further brings 1.3\% and 2.5\% increase in T2I R@1 and I2T R@1 on MSCOCO. Generally, we notice that the performance converges when $K \in (6, 10)$. However, increasing $K$ introduces extra computation overhead, since the text encoder process $K$ captions in every iteration. %
Therefore, we choose $K=8$ as our main setting, as it achieves a good balance between optimal performance and computation.

\begin{table}[t]
	\centering\scriptsize
	\SetTblrInner{rowsep=1.2pt}
	\SetTblrInner{colsep=2pt}
    \resizebox{\columnwidth}{!}{
	\begin{tblr}{
		cells={halign=c,valign=m},  
        hline{1,10}={1.0pt},
        hline{3}={},
        hline{5}={},
        hline{2}={2-3}{leftpos = -1, rightpos = -1, endpos},
        hline{2}={4-5}{leftpos = -1, rightpos = -1, endpos},
        hline{2}={6-7}{leftpos = -1, rightpos = -1, endpos},
        hline{2}={8-8}{leftpos = -1, rightpos = -1, endpos},
        hline{2}={9-9}{leftpos = -1, rightpos = -1, endpos},
        cell{1}{2}={c=2}{},
        cell{1}{4}={c=2}{},
        cell{1}{6}={c=2}{},
        }
    & MSCOCO & & DOCCI & & Urban-1K & & VOC20 & ImageNet \\
    \bf K & T2I@1 & I2T@1 & T2I@1 & I2T@1 & T2I@1 & I2T@1 & mIOU & Top-1 \\
    CLIP~\cite{cherti2023reproducible}                        &    27.0   &   38.9    &    10.3   &    25.0   &    41.3   &  37.7     &    3.16   &   23.8    \\
    SigLIP~\cite{zhai2023sigmoid}                        &     28.3  &  40.1     &    10.4   &   24.9    &    42.8  &    40.5   &   3.1   &   25.4   \\
    2       &    36.4   &    49.1   &   13.9    &    35.5   &   68.7    &   62.9    &    56.3   &   31.2    \\
    4       &    36.7   &    49.8   &   14.2    &   35.4    &    69.1   &    62.7   &   57.4    &   32.8    \\
    6                      &    37.4   &  51.2     &   14.9    &   35.4    &  69.8     &       61.4 &  59.5  &   33.6    \\
    8      &    37.7   &   51.6    &    \textbf{15.1}   &   \textbf{35.7}    &    69.5   &   63.5    &    59.7   &    \textbf{33.8}  \\
    10     &    \textbf{37.8}   &   \textbf{51.7}   &  15.0    &  35.1     &  \textbf{71.6}    &   \textbf{64.2}    &   \textbf{60.9}   &    33.6   \\
    \end{tblr}
    }
    \caption{Ablation results on the number of sub-captions $K$ for \method. OpenCLIP~\cite{cherti2023reproducible}, SigLIP~\cite{zhai2023sigmoid} and \method are pre-trained on CC3M-recap datasets under the same configuration. All models use ViT-B/16 as vision encoder. The best results are \textbf{bold}.}
    \label{tab:k_ablation}
\end{table}

\section{Implementation Details}
\label{sec:implementation_details}

\begin{table}[t]
\centering
\scriptsize
\SetTblrInner{rowsep=1.2pt}
\SetTblrInner{colsep=2pt}
\resizebox{\linewidth}{!}{
\begin{tblr}{
    cells={halign=c,valign=m},
    column{1}={halign=l},
    hline{1,11}={1.0pt},
    hline{2}={},
    cell{3}{2}={c=4}{},
    cell{4}{2}={c=4}{},
    cell{6}{2}={c=4}{},
    cell{8}{2}={c=4}{},
    cell{9}{2}={c=4}{},
    cell{10}{2}={c=4}{},
}
Config & CC3M-recap & CC12M-recap & YFCC15M-recap & Merged-30M \\
Batch size & $1,024$ & $6.134$ & $6,134$ & $6,134$ \\
Optimizer & AdamW~\cite{loshchilov2017decoupled} & & & \\
Learning rate & $5\times10^{-4}$ & & & \\
Weight decay & $0.5$ & $0.5$ & $0.5$ & $0.2$ \\
Adam $\beta$ & $\beta_1, \beta_2=(0.9, 0.98)$ & & & \\
Adam $\epsilon$ & $1\times10^{-8}$ & $1\times10^{-8}$ & $1\times10^{-8}$ & $1\times10^{-6}$ \\
Total epochs & $32$ & & & \\
Warm up & $2,000 \text{(steps)}$ & & & \\
LR scheduler & cosine decay & & & \\
\end{tblr}
}
\caption{Pre-training hyper-parameters for \method and all re-trained baseline methods. LR scheduler: Learning Rate scheduler.}
\label{tab:hyperparam}
\end{table}

\begin{table}[t]
\centering
\scriptsize
\SetTblrInner{rowsep=1.2pt}
\SetTblrInner{colsep=2pt}
\resizebox{.8\linewidth}{!}{
\begin{tblr}{
    cells={halign=c,valign=m},  
    hline{1,13}={1.0pt},       
    hline{2,5,7,9,11}={},        
    cell{5}{2}={r=2}{},
    cell{7}{2}={r=2}{},
    cell{9}{2}={r=2}{},
    cell{11}{2}={r=2}{},
}
    Method & 
    \rotatebox[origin=lb]{90}{Data Size} & 
    \rotatebox[origin=lb]{90}{VOC20} & 
    \rotatebox[origin=lb]{90}{Cityscapes} & 
    \rotatebox[origin=lb]{90}{Context59} & 
    \rotatebox[origin=lb]{90}{ADE20K} & 
    \rotatebox[origin=lb]{90}{COCO-Stuff } & 
    \rotatebox[origin=lb]{90}{Average} \\
    
    CLIP~\cite{radford2021learning}        & 400M & 41.8  & \textbf{5.5} & \textbf{ 9.2 } &  \textbf{3.2 }& 4.4  &  12.8 \\
    OpenCLIP~\cite{cherti2023reproducible} & 2B &  \textbf{47.2} &  5.1 &  9.0  &  2.9 & \textbf{5.0}  &  \textbf{13.9} \\
    MetaCLIP~\cite{xu2023demystifying}     & 2.5B & 35.4  &  5.0 & 8.1  &  2.2 & 4.3  & 11.0 \\
    FLAIR-\textit{CLIP} & 3M & \textbf{60.9} & 8.9 & 15.6 & 8.0 & 9.7 &  20.6  \\
    FLAIR-\textit{TC} &  & 53.9 & \textbf{20.6} & \textbf{23.8} & \textbf{13.1 }& \textbf{13.1} & \textbf{24.9} \\
    FLAIR-\textit{CLIP} & 12M & \textbf{69.7}& 14.5 & 17.4 & 10.0 & 12.2 & 24.8 \\
    FLAIR-\textit{TC} &  & 55.1 & \textbf{20.1}& \textbf{22.9} & \textbf{13.3} & \textbf{15.4} &  \textbf{25.4} \\
    FLAIR-\textit{CLIP} & 15M & \textbf{71.5} & 13.3 & \textbf{18.4} & 9.0 & 12.5 & \textbf{24.9} \\
    FLAIR-\textit{TC} &  & 49.2 & \textbf{16.5} & 17.4 & \textbf{9.1} & \textbf{13.6} &  21.2\\
    FLAIR-\textit{CLIP} & 30M & \textbf{73.0} & \textbf{13.6} & \textbf{18.6} &  10.4& 13.3 & \textbf{25.8} \\
    FLAIR-\textit{TC} &    &  48.3 & \textbf{13.6} & 17.4 & \textbf{10.8} & \textbf{14.4} &   20.9 \\
\end{tblr}
}
\caption{Mean Intersection over Union (mIoU) for zero-shot semantic segmentation on VOC20~\cite{everingham2015pascal}, Cityscapes~\cite{cordts2016cityscapes}, Context59~\cite{mottaghi2014role}, ADE20K~\cite{zhou2019semantic}, and COCO-Stuff~\cite{caesar2018coco}. All models employ ViT-B/16 as the vision encoder. The best results are \textbf{bold}.
}
\label{tab:ss_supplementary}
\end{table}

In this section, we  describe the detailed implementation of pre-training and downstream tasks evaluation. 

\noindent
\myparagraph{Pre-training.} 
Our implementation is based on the OpenCLIP~\cite{cherti2023reproducible} code base with the ViT-B/16 architecture for the image encoder. Both image and text encoder consist of 12 transformer layers, and the embedding size is fixed at 512. Specifically for \method, we replace the final pooling layer of image encoder with our \pooling, while the rest of the layers remain unchanged. Our loss function initializes \( t \) at 0.07 and \( b \) at -10, consistent with the settings used in SigLIP.
We follow DreamLIP's pre-training configuration as displayed in \cref{tab:hyperparam}. However, we use 6K batch size for CC12M-recap, YFCC15M-recap and Merged30M due to GPU RAM limit. Experiments on CC3M-recap are trained on 8 NVIDIA A100 40GB GPUs and 32 GPUs on the other datasets. All baseline models, CLIP and SigLIP, follow the same pre-training configurations.

\noindent
\myparagraph{Large-scale Pre-trained CLIP Models.}
In the main paper, we report the values for OpenCLIP (2B) and SigLIP (10B). Both models employ ViT-B/16 as the vision encoder. Those values were obtained by evaluating the pre-trained weights of OpenCLIP. ``OpenCLIP (2B)'' refers to the ViT-B/16 model trained on the LAION-2B dataset with the pre-trained name of ``laion2b\_s34b\_b88k''. ``SigLIP (10B)'' refers to the ViT-B/16-SigLIP model trained on the WebLI dataset with the pre-trained name of ``webli''. The Llip~\cite{lavoiemodeling} and MetaCLIP~\cite{xu2023demystifying} results for zero-shot image classification are directly obtained from their papers.

\noindent
\myparagraph{Zero-shot Semantic Segmentation.}
\label{sec:detailed_seg}
As discussed in Sec.~4 of the main paper, zero-shot semantic segmentation is based on the similarity between local image tokens and global text queries \(\left\{ \langle \mathbf{v}^{\text{loc}}_{i}, \mathbf{t}^{\text{g}}_{j} \rangle \mid j \in \{1, 2, \dots, M\} \right\}\), where \(M\) represents the number of class names in the dataset. Compared to CLIP, a key advantage of \method is its flexibility during inference: it can either directly compute \(\langle \mathbf{v}^{\text{loc}}_{i}, \mathbf{t}^{\text{g}}_{j} \rangle\) without applying \(f_{\text{AttnPool}}(.)\) (\method-\textit{CLIP}), or first use \(f_{\text{AttnPool}}(\mathbf{v}^{\text{loc}}_{i}, \mathbf{t}^{\text{g}}_{j})\) to generate fine-grained embeddings \(\mathbf{v}^{\text{tc}}_{i,j}\), and then compute \(\langle \mathbf{v}^{\text{tc}}_{i,j}, \mathbf{t}^{\text{g}}_{j} \rangle\) (\method-\textit{TC}). Segmentation results for both approaches are reported in \cref{tab:ss_supplementary}. For implementation details, including window size, stride, and other parameters, we follow the design choices described in~\cite{wang2023sclip}.

Interestingly, using the CLIP method increases mIoU on VOC20 by approximately 10\%, while the TC method improves performance on other datasets. Both methods outperform OpenCLIP and SigLIP models trained on billions of images. This indicates that the segmentation capability of \method is not solely reliant on the attention pooling layer, because the local image tokens \(\mathbf{v}_{\text{loc}}\) encode strong localization information independently.

\noindent
\myparagraph{Zero-shot Image Classification.}  
We follow the prompt ensemble strategy described in LaCLIP~\cite{fan2024improving} and ALIP~\cite{yang2023alip}, employing the same prompt templates. For each class name, we compute the average text embedding across all templates, which is then used to calculate the similarity between test images and class embeddings. For zero-shot ImageNet classification, we use the seven prompt templates recommended by \cite{radford2021learning}, consistent with LaCLIP~\cite{fan2024improving}.

\end{document}